%% file: neurips_2026.tex
\documentclass{article}
\usepackage[T1]{fontenc}
\usepackage[utf8]{inputenc}
\usepackage{textcomp}

\PassOptionsToPackage{table}{xcolor}

\usepackage{microtype}
\usepackage{graphicx}
\usepackage{subcaption}
\usepackage{booktabs} 
\usepackage{wasysym}  

\usepackage{hyperref}
\hypersetup{hidelinks}

\usepackage[preprint]{neurips_2026}

\usepackage{amsmath}
\usepackage{amssymb}
\usepackage{mathtools}
\usepackage{amsthm}

\usepackage{xcolor}
\usepackage{tikz}
\usepackage{wrapfig}
\usepackage{float}
\usepackage{placeins}
\usepackage{listings}
\usepackage{comment}
\usepackage{colortbl}
\providecommand{\NewStructureName}[1]{}
\providecommand{\AssignStructureRole}[2]{}
\providecommand{\NewTaggingSocket}[2]{}
\providecommand{\NewTaggingSocketPlug}[3]{}
\providecommand{\AssignTaggingSocketPlug}[2]{}
\providecommand{\UseTaggingSocket}[1]{}

\providecommand{\tagstructbegin}[1]{}

\providecommand{\tagpdfparaOff}[1]{}
\usepackage{tcolorbox}
\tcbuselibrary{skins,breakable}
\usepackage{enumitem}
\usepackage{array}
\usepackage{tabularx}
\usepackage{etoc}
\usepackage[toc]{appendix} 
\usepackage{listingsutf8} 
\usetikzlibrary{positioning,shapes,fit, patterns}
\usetikzlibrary{shapes.geometric, arrows, positioning, fit, decorations.pathreplacing, calc}

\lstdefinestyle{verbatimish}{
  inputencoding=utf8,
  basicstyle=\ttfamily\small,
  columns=fullflexible,
  upquote=true,
  keepspaces=true,
  showstringspaces=false,
  breaklines=true,
  escapechar=§,
}
\lstdefinestyle{codeblockstyle}{
  inputencoding=utf8,
  basicstyle=\ttfamily\small,
  columns=fullflexible,
  upquote=true,
  keepspaces=true,
  showstringspaces=false,
  breaklines=true,
  escapechar=§,
  frame=single,
  rulecolor=\color{black!25},
  framerule=0.4pt,
  framesep=4pt,
  backgroundcolor=\color{gray!5},
}
\lstnewenvironment{codeblock}[1][]%
  {\lstset{style=codeblockstyle,#1}}%
  {}

\definecolor{darkheader}{RGB}{70,70,70}
\definecolor{lightgray}{RGB}{245,245,245}
\definecolor{stage1color}{RGB}{135, 206, 235}
\definecolor{stage2color}{RGB}{255, 182, 193}
\definecolor{stage3color}{RGB}{144, 238, 144}
\definecolor{inputcolor}{RGB}{255, 255, 224}
\definecolor{outputcolor}{RGB}{216, 191, 216}
\definecolor{codeblue}{RGB}{40,80,120}
\definecolor{codered}{RGB}{180,40,40}
\definecolor{codegreen}{RGB}{60,120,60}
\definecolor{codegray}{RGB}{100,100,100}
\definecolor{codepurple}{RGB}{120,40,120}
\definecolor{codeorange}{RGB}{200,100,20}
\definecolor{codemagenta}{RGB}{150,40,150}
\definecolor{codebg}{RGB}{245,245,245}

\newtcolorbox{promptbox}[2][]{
    colback=lightgray,
    colframe=black,
    coltitle=white,
    colbacktitle=darkheader,
    fonttitle=\bfseries\large,
    title=#2,
    boxrule=1pt,
    titlerule=0pt,
    toptitle=8pt,
    bottomtitle=8pt,
    #1
}

\lstdefinestyle{colorfulstyle}{
  basicstyle=\ttfamily\footnotesize,
  keywordstyle=\color{codeblue}\bfseries,
  stringstyle=\color{codered},
  commentstyle=\color{codegreen},
  numberstyle=\tiny\color{codegray},
  numbers=left,
  numbersep=5pt,
  breaklines=true,
  breakatwhitespace=true,
  breakautoindent=true,
  postbreak=\mbox{\textcolor{codegray}{$\hookrightarrow$}\space},
  captionpos=b,
  keepspaces=true,
  showspaces=false,
  showstringspaces=false,
  showtabs=false,
  tabsize=2,
  frame=single,
  backgroundcolor=\color{gray!5},
  literate=
    {α}{{$\alpha$}}1
    {β}{{$\beta$}}1
    {∀}{{$\forall$}}1
    {∃}{{$\exists$}}1
    {≤}{{$\le$}}1
    {≥}{{$\ge$}}1
    {→}{{$\to$}}1
    {λ}{{$\lambda$}}1
    {∧}{{$\land$}}1
    {∨}{{$\lor$}}1
    {¬}{{$\neg$}}1
    {…}{{\ldots}}1
    {⋯}{{\ldots}}1
    {·}{{$\cdot$}}1
    {∈}{{$\in$}}1
    {×}{{$\times$}}1
    {≠}{{$\neq$}}1
}

\lstdefinestyle{promptstyle}{
    basicstyle=\ttfamily\footnotesize\bfseries,
    keywordstyle=\color{blue}\bfseries,
    stringstyle=\color{red}\bfseries,
    commentstyle=\color{green!60!black}\bfseries,
    numberstyle=\tiny\color{gray}\bfseries,
    breakatwhitespace=false,
    breaklines=true,
    captionpos=b,
    keepspaces=true,
    numbers=left,
    numbersep=5pt,
    showspaces=false,
    showstringspaces=false,
    showtabs=false,
    tabsize=2,
    frame=single,
    frameround=tttt,
    backgroundcolor=\color{gray!10},
    rulecolor=\color{black!30},
    emphstyle=\color{purple}\bfseries,
    moredelim=[s][\color{orange}\bfseries]{\{\{} {\}\}},
    moredelim=[s][\color{magenta}\bfseries]{<}{>},
}

\lstdefinelanguage{Lean}{
  keywords={def, theorem, lemma, import, by, sorry, match, with, let, in, if, then, else, fun, λ, structure, inductive},
  keywordstyle=\color{codeblue}\bfseries,
  sensitive=true,
  comment=[l]{--},
  commentstyle=\color{codegreen},
  string=[b]",
  stringstyle=\color{codered},
}

\newtcolorbox{promptboxICML}[1]{
  title=#1,
  breakable,
  enhanced,
  colback=lightgray,
  colframe=black,
  coltitle=white,
  colbacktitle=darkheader,
  fonttitle=\bfseries\small,
  boxrule=0.8pt,
  left=4pt,
  right=4pt,
  top=4pt,
  bottom=4pt,
  before skip=2pt,
  after skip=2pt
}

\usepackage[capitalize,noabbrev]{cleveref}

\theoremstyle{plain}

\theoremstyle{definition}

\theoremstyle{remark}

\usepackage[textsize=tiny]{todonotes}

\title{BRIDGE: Building Representations in Domain-Guided Verified Program Synthesis}

\author{%
  Robert Joseph George \\
  California Institute of Technology \\
  Amazon \\
  \texttt{rgeorge@caltech.edu} \\
  \And
  Carson Eisenach \\
  Amazon \\
  \And
  Udaya Ghai \\
  Amazon \\
  \And
  Dominique Perrault-Joncas \\
  Amazon \\
  \And
  Anima Anandkumar \\
  California Institute of Technology \\
  \And
  Dean Foster \\
  Amazon \\
}

\begin{document}

\maketitle

\begin{abstract}
Large language models (LLMs) are increasingly used to write code, but generated programs can look correct while still containing subtle bugs, missing edge cases, or mismatched assumptions. This motivates \emph{verifiable coding}, where the output includes not only code but also formal artifacts that justify the code against a specification. In proof assistants such as \textsc{Lean}, this is a multi artifact prediction problem: executable code, specifications, theorem statements, and proof attempts must remain mutually consistent. We present \textbf{BRIDGE}, a structured prompting framework that decomposes this process into \emph{Code}, \emph{Specification}, and \emph{Theorem/Proof} domains. BRIDGE uses reasoning tailored to each domain, together with checks across artifacts, to reduce semantic drift. Our primary code metric is \textsc{Lean} executable correctness: the generated program must elaborate, satisfy Lean's termination and totality checks, and pass benchmark tests. This metric is stricter than ordinary unit test accuracy, but it is not a substitute for full semantic verification. Across our 178 problem Lean benchmark and the public \textsc{VERINA} and \textsc{CLEVER} benchmarks, BRIDGE improves executable correctness by up to $1.5\times$ and reaches comparable success with up to roughly $2\times$ fewer Lean evaluations. Supervised fine tuning on BRIDGE style functional traces improves over matched direct trace fine tuning, suggesting that the representation can be internalized rather than used only as a prompt. BRIDGE also improves specification guided code generation and downstream theorem/proof diagnostics, producing more Lean elaborating theorem statements and more completed proof attempts aligned to the implementation. BRIDGE\footnote{Code and data are available at \url{https://github.com/lean-dojo/BRIDGE}.} provides a verification oriented inductive bias for connecting code, specifications, theorem statements, and proof attempts, while leaving full semantic proof completion at scale as an open problem.
\end{abstract}

\section{Introduction}

Program verification aims for guarantees stronger than passing tests: an implementation should satisfy a precise specification for \emph{all} valid inputs and executions~\cite{hoare1969axiomatic,floyd1967assigning}. In modern proof assistants, this goal requires a collection of coupled artifacts rather than a single generated object: executable code, a formal specification, theorem statements connecting the code to the specification, and proof scripts checked by the kernel~\cite{demoura2015lean}. Treating these artifacts independently makes semantic drift easy. A model can emit code that passes tests, a specification that typechecks but is vacuous, a theorem statement that looks correct but fails to capture the intended property, or a proof of a weak claim that says little about the original program. Current LLM code generation systems struggle in this setting because Lean requires more than an algorithm sketch~\cite{chen2021codex,li2022competition}. Recursive calls must be justified, types must agree with the intended invariants, and specifications must be strong enough to make the theorem meaningful without becoming impossible to prove~\cite{demoura2015lean,mathlib2020}. Verification therefore exposes a representation problem: the model must translate informal intent into a sequence of artifacts whose syntax, types, recursion structure, contracts, and proof obligations agree with one another.

\begin{figure}
    \centering
    \includegraphics[width=1.0\linewidth]{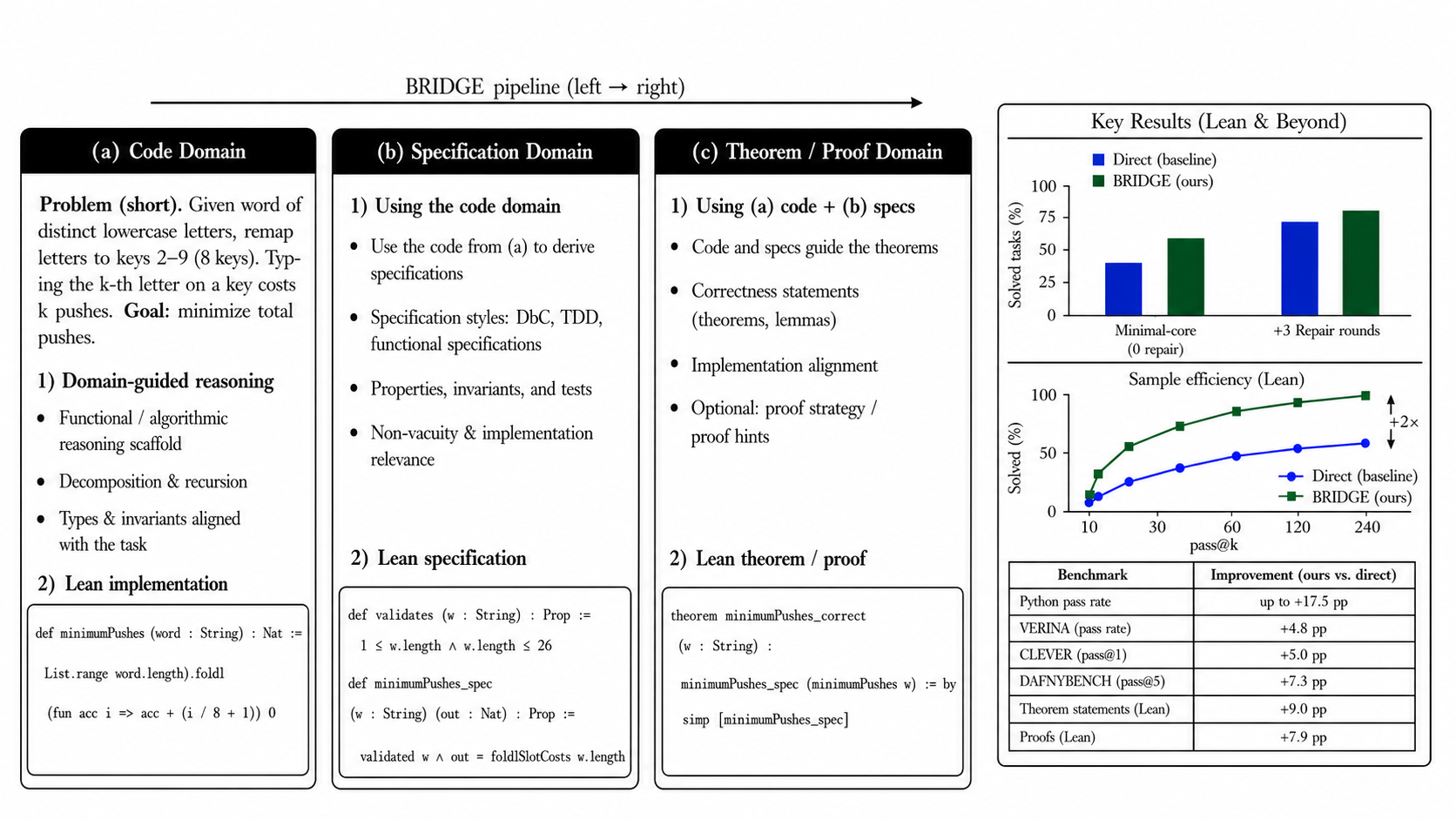}
\caption{\textbf{BRIDGE methodology and key results.} BRIDGE decomposes verifiable coding into linked Code, Specification, and Theorem/Proof domains. The example shows how a problem statement is converted into functional reasoning, specification reasoning, and theorem/proof reasoning, while the accompanying results summarize the main empirical gains from functional scaffolding. Appendix~\ref{app:prompt_system} gives the detailed prompt templates used to instantiate these domains.}
\label{fig:bridge-min-key-pushes}
\end{figure}

Interactive theorem provers such as \textsc{Lean} make this coupling especially sharp because Lean unifies programming, specification, and proof inside a single dependent type theory. Generated programs must elaborate, satisfy typing constraints, and respect Lean's termination and totality discipline~\cite{demoura2015lean,mathlib2020}. We therefore use \emph{Lean executable correctness} as a scalable code metric: a solution must elaborate, satisfy termination and totality checks, and pass the benchmark tests. Full formal verification remains the long term goal. We propose \textbf{BRIDGE}, a structured prompting framework for generating and checking linked verification artifacts. BRIDGE fixes a three domain decomposition, \textbf{Code}, \textbf{Specifications}, and \textbf{Theorem/Proof Artifacts}, but allows flexible ordering and conditioning between them. In Lean, we find an ordering that begins with code especially effective: the model first generates Lean compatible code using an aligned scaffold, then uses that implementation as a semantic anchor for specifications, theorem statements, and targeted proof attempts.

BRIDGE is not a verified compiler from \textsc{OCaml}, \textsc{Haskell}, or \textsc{Python} into Lean. The intermediate program is a reasoning scaffold, not a certified source artifact. Semantic confidence instead comes from tests, elaboration, termination checks, nonvacuity checks, implementation relevance, proof attempts, and human inspection. BRIDGE studies the front end of verified synthesis, not arbitrary proof completion at scale. Its purpose is to make the code, specification, theorem, and proof attempt pipeline easier to inspect and easier for downstream proof search to use.

Our main empirical finding is that \emph{reasoning structure matters}. On our 178 problem Lean benchmark, functionally aligned intermediate reasoning, using \textsc{OCaml} or \textsc{Haskell} style scaffolds, improves executable Lean synthesis over direct prompting and over Double Lean, a length controlled baseline where the model performs an extra Lean style reasoning step before producing final Lean code. In the minimal core setting, functional reasoning solves 59/178 tasks, compared with 44/178 for direct prompting and 49/178 for Double Lean. With three repair rounds, the gap narrows but remains positive, with 83/178 solved by functional reasoning versus 77/178 for direct prompting and 81/178 for Double Lean. We also find that specification oriented prompting improves conventional code generation, that BRIDGE style traces can be internalized through supervised fine tuning, and that functional scaffolds produce Lean code that is easier to attach theorem statements and proof attempts to. To improve comparability, we additionally evaluate on public benchmarks, including \textsc{VERINA}, \textsc{CLEVER}, and a solver backed \textsc{DafnyBench} pilot. These runs show functional gains across several independent settings, while also clarifying that model family, prompt calibration, and search budget still matter.

\textbf{Contributions.}
\begin{enumerate}[leftmargin=1.2em]
    \item \textbf{A multi artifact framework for verifiable coding.}
    We introduce \textbf{BRIDGE}, which decomposes verifiable coding into linked Code, Specification, and Theorem/Proof domains and uses reasoning over artifacts to reduce semantic drift between them.

    \item \textbf{Functional reasoning improves Lean executable synthesis.}
    On our 178 problem Lean benchmark, functionally aligned intermediate reasoning improves executable correctness by up to $1.5\times$ over direct prompting and reaches comparable success with up to roughly $2\times$ fewer Lean evaluations. A length controlled Double Lean baseline shows that the gains are not explained by longer generations alone.

    \item \textbf{BRIDGE improves specifications and theorem/proof artifacts.}
    Specification guided prompting improves code generation by up to 17.5 percentage points and produces more useful formal artifacts. Functional scaffolds yield more Lean elaborating theorem statements and more completed proof attempts aligned to the implementation, suggesting that better code structure also makes downstream verification artifacts easier to attach.

    \item \textbf{The gains transfer and can be learned.}
    BRIDGE improves public Lean benchmarks such as \textsc{VERINA} and \textsc{CLEVER}, as well as code verification with solver backends such as \textsc{DafnyBench}. Supervised fine tuning on functional BRIDGE traces improves over matched direct trace fine tuning, most clearly on \textsc{VERINA} (20.3\% to 29.4\%, a 9.1 percentage point gain), suggesting that BRIDGE captures a learnable inductive bias rather than a one time prompt effect.
\end{enumerate}

\begin{figure*}[!t]
    \centering
\begin{subfigure}[t]{0.49\linewidth}
    \centering
    \includegraphics[width=\linewidth,height=0.30\textheight,keepaspectratio]{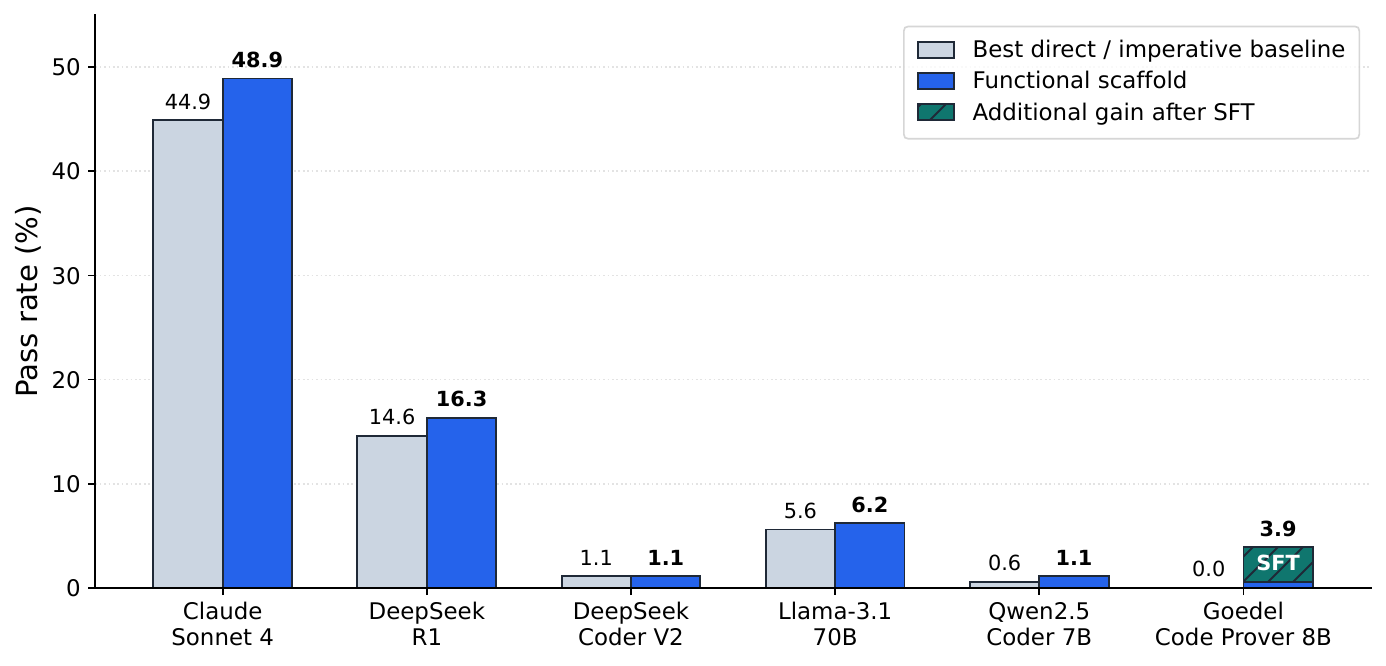}
    \caption{Functional gains across models, including Goedel SFT.}
\end{subfigure}
\hfill
\begin{subfigure}[t]{0.49\linewidth}
    \centering
    \includegraphics[width=\linewidth,height=0.30\textheight,keepaspectratio]{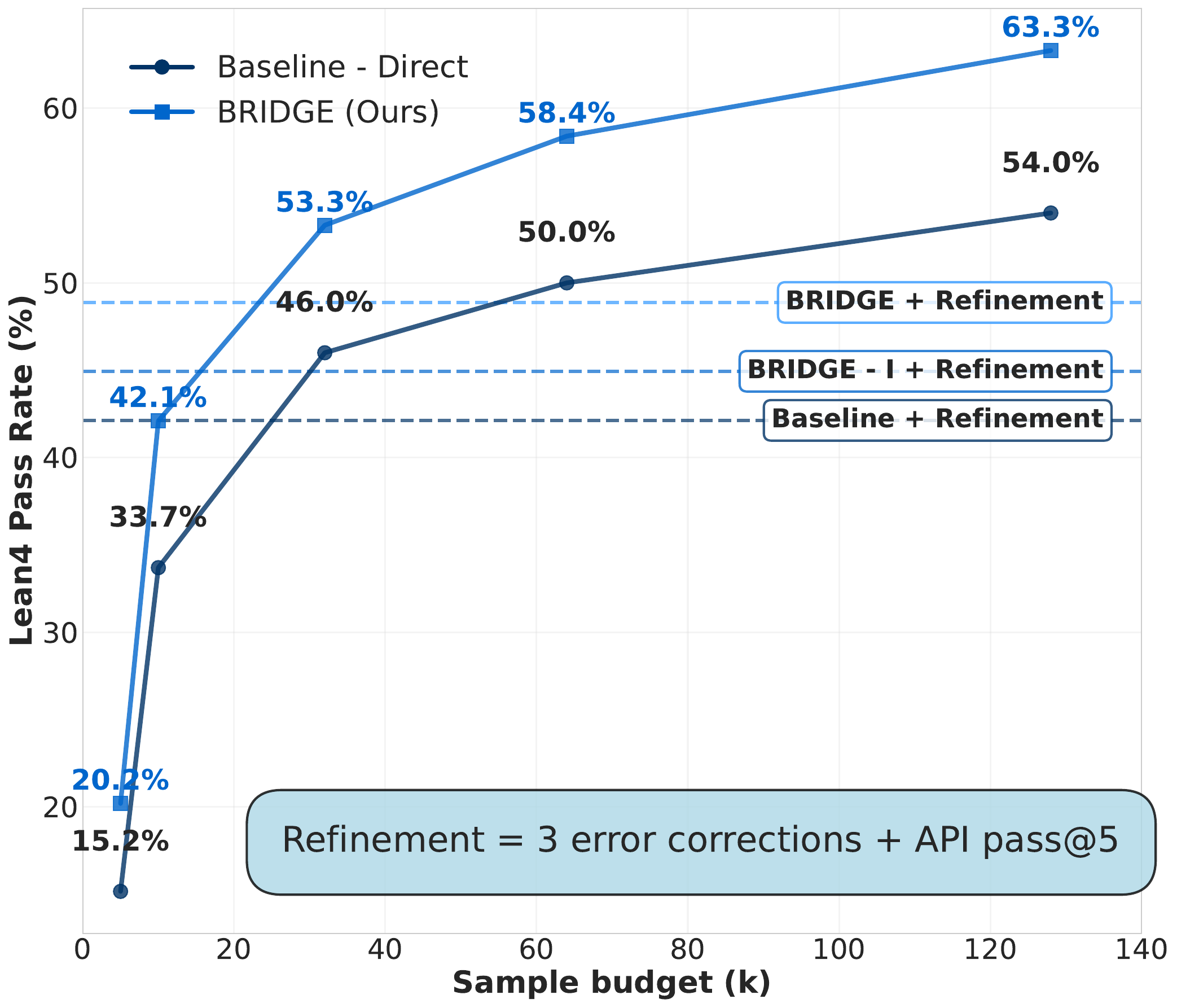}
    \caption{Sample efficiency versus generation length.}
\end{subfigure}
\caption{BRIDGE suite results. (a) Functional scaffolds improve executable Lean synthesis across models, including the Goedel SFT setting. (b) Functional scaffolds reach comparable success with fewer Lean evaluations, even when compared with length controlled baselines.}
    \label{fig:bridge178_main_results}
\end{figure*}

Though instantiated in \textsc{Lean}, BRIDGE is compatible with other verification settings such as \textsc{rocq}, \textsc{Dafny}, and \textsc{Boogie}, where code, specifications, and correctness claims must be kept aligned.

\vspace{-0.3cm}
\section{Related Work}
\label{sec:related}

Formal verification spans verification with solver backends, interactive theorem proving, and program synthesis under formal specifications. Systems such as \textsc{Dafny}, \textsc{Boogie}, and \textsc{Verus} translate contracts and annotations into verification conditions that SMT solvers can often discharge automatically~\cite{loughridge2024dafnybench,zhu2023verus}. Interactive theorem proving is more demanding because programs, specifications, theorem statements, and proofs must remain aligned inside a richer formal environment. Programming paradigm also matters: imperative languages rely on mutation, loops, and state transitions, while functional languages emphasize immutable data, pattern matching, and structural recursion, which align more naturally with proof assistant programming~\cite{hoare1971proof,hudak1989conception}. BRIDGE targets this gap by using intermediate reasoning structures that better match the target verification system.

Recent LLM verification systems show both promise and brittleness. Public Lean benchmarks such as \textsc{VERINA} and \textsc{CLEVER} show that models can generate well formed code, specifications, theorem statements, and proof attempts, but still struggle to connect them into reliable verified artifacts~\cite{ye2025verina,clever2025}. Solver based systems such as \textsc{Clover} and \textsc{AutoSpec}, and benchmarks such as \textsc{DafnyBench}, show stronger results when generated annotations are close enough for automated backends to finish the remaining proof obligations~\cite{sun2024clover,wen2024autospec,loughridge2024dafnybench}. Specification synthesis systems such as \textsc{VeriSpecGen} improve Lean specifications through requirement decomposition and localized repair~\cite{ye2026verispecgen}. Complementary Lean infrastructure such as \textsc{TorchLean} studies a different part of the verification stack by giving neural network programs a shared formal semantics for execution and verification inside Lean~\cite{george2026torchlean}. BRIDGE is orthogonal to these systems: it studies how reasoning representation and artifact coupling affect code, specification, theorem, and proof generation. Table~\ref{tab:cot_software_verification} summarizes the closest methods and benchmarks; Appendix~\ref{app:related_work} gives extended related work.

\begin{table*}[!t]
\centering
\caption{Reasoning methods and verification benchmarks. $\bullet$ = supported or evaluated; $\circ$ = not a focus; $\LEFTcircle$ = theorem statements only; $\RIGHTcircle$ = proof attempts or proof evaluation only. ``Compose'' indicates whether the method is designed to connect code, specifications, theorem statements, and proof attempts.}
\label{tab:cot_software_verification}
\renewcommand{\arraystretch}{1.12}
\setlength{\tabcolsep}{3pt}
\small
\resizebox{\textwidth}{!}{%
\begin{tabular}{lccccc}
\toprule
\textbf{Work / Method} & \textbf{Reasoning Style} & \textbf{Code} & \textbf{Spec} & \textbf{Theorem/Proof} & \textbf{Compose} \\
\midrule
\rowcolor{gray!15}
\multicolumn{6}{l}{\textbf{Prompting and reasoning methods}} \\
\textsc{CoT Prompting}~\cite{wei2022cot} & Rationale & $\bullet$ & $\circ$ & $\circ$ & $\circ$ \\
\textsc{Self-Consistency}~\cite{wang2022selfconsistency} & Sample and vote & $\bullet$ & $\circ$ & $\circ$ & $\circ$ \\
\textsc{PAL}~\cite{gao2022pal} & Program aided & $\bullet$ & $\circ$ & $\circ$ & $\circ$ \\
\textsc{ReAct}~\cite{yao2023react} & Reason and act & $\bullet$ & $\circ$ & $\circ$ & $\circ$ \\
\textsc{Tree-of-Thought}~\cite{yao2023tot} & Search & $\bullet$ & $\circ$ & $\circ$ & $\circ$ \\
\textsc{Reflexion}~\cite{shinn2023reflexion} & Self reflection & $\bullet$ & $\circ$ & $\circ$ & $\circ$ \\
\textsc{Self-Refine}~\cite{madaan2023selfrefine} & Iterative critique & $\bullet$ & $\circ$ & $\circ$ & $\circ$ \\
\textsc{Clover}~\cite{sun2024clover} & Closed loop checks & $\bullet$ & $\bullet$ & $\circ$ & $\circ$ \\
\textsc{AutoSpec}~\cite{wen2024autospec} & Spec synthesis & $\circ$ & $\bullet$ & $\circ$ & $\circ$ \\
\rowcolor{orange!15}
\textbf{BRIDGE (ours)} & Domain guided & $\bullet$ & $\bullet$ & $\bullet$ & $\bullet$ \\
\midrule
\rowcolor{gray!15}
\multicolumn{6}{l}{\textbf{Benchmarks}} \\
\textsc{DafnyBench}~\cite{loughridge2024dafnybench} & Solver backed & $\circ$ & $\bullet$ & $\circ$ & $\circ$ \\
\textsc{miniCodeProps}~\cite{lohn2024minicodeprops} & Lean properties & $\circ$ & $\circ$ & $\RIGHTcircle$ & $\circ$ \\
\textsc{VERINA}~\cite{ye2025verina} & Lean artifacts & $\bullet$ & $\bullet$ & $\RIGHTcircle$ & $\circ$ \\
\textsc{CLEVER}~\cite{clever2025} & Lean verification & $\bullet$ & $\bullet$ & $\RIGHTcircle$ & $\circ$ \\
\textsc{DAFNYCOMP}~\cite{dafnycomp2025} & Compositional specs & $\circ$ & $\bullet$ & $\circ$ & $\circ$ \\
\textsc{Vericoding}~\cite{vericoding2025} & Cross system & $\bullet$ & $\bullet$ & $\RIGHTcircle$ & $\circ$ \\
\bottomrule
\end{tabular}
}
\end{table*}
\vspace{-0.5cm}
\section{Methodology}
\label{sec:methodology}
\vspace{-0.2cm}
\textbf{BRIDGE} treats verifiable coding as an artifact alignment problem rather than a single code generation task. It organizes generation into three linked domains: \emph{Code} for executable implementations, \emph{Specifications} for formal behavioral constraints, and \emph{Theorem/Proof Artifacts} for correctness claims and bounded proof attempts. 
In our Lean setting, we often use an ordering that begins with code because current models anchor more reliably on executable implementations than on fully faithful semantic specifications, though the same framework also supports specification first and theorem first variants. No artifact is validated in isolation: code may compile with the wrong behavior, specifications may be vacuous, theorem statements may miss the intended claim, and proofs may establish weak properties. We therefore evaluate artifacts through checks across artifacts, including elaboration, termination and totality acceptance, tests, nonvacuity, implementation relevance, wrong implementation rejection, and proof completion without \texttt{sorry} where applicable. Full prompt templates are in Appendix~\ref{app:prompt_system}.

\subsection{The Code Domain}
\label{sec:code_domain}

The Code domain produces the final \textsc{Lean} implementation. Lean is both a programming language and a proof assistant, so generated programs must elaborate, typecheck, and satisfy Lean's termination and totality discipline~\cite{demoura2015lean,mathlib2020}. Recursive definitions often require structural recursion or a well founded argument, and implementation choices can determine whether later specifications and theorem statements are even expressible. BRIDGE compares direct Lean generation with intermediate reasoning scaffolds. Functional scaffolds, inspired by \textsc{OCaml} and \textsc{Haskell}, encourage immutable data, pattern matching, recursion, and compositional definitions. Imperative scaffolds, inspired by \textsc{Python} and \textsc{C++}, encourage loops, mutation, arrays, and state transitions. This comparison reflects a classical distinction in program reasoning: imperative programs often require explicit invariants over changing state, while functional programs more naturally support inductive reasoning over recursive structure~\cite{hoare1971proof,hudak1989conception}. The intermediate program is only a scaffold. It is not required to compile in its source language, and BRIDGE is not a verified compiler from \textsc{OCaml}, \textsc{Haskell}, \textsc{Python}, or \textsc{C++} into Lean. All code results are evaluated only on the final Lean output.

\subsection{The Specification Domain}
\label{sec:spec_domain}

The Specification domain describes what the implementation is intended to guarantee. In Lean, specifications can appear as predicates over inputs and outputs, preconditions, postconditions, invariants, dependent types, or auxiliary definitions that characterize valid behavior. Following the classical view that correctness means satisfying a specification for all admissible inputs~\cite{hoare1969axiomatic,floyd1967assigning,dijkstra1976discipline}, BRIDGE uses specification reasoning to make assumptions, edge cases, and intended guarantees explicit. We instantiate this domain through several reasoning styles, including Design by Contract, Property Based Testing, Test Driven Development, and Defensive Programming. These styles provide independent views of the same task: contracts expose preconditions and postconditions, property reasoning exposes universal behavior such as bounds or monotonicity, test driven reasoning exposes edge cases, and defensive reasoning exposes invalid inputs or hidden assumptions. A specification that compiles is not necessarily meaningful. It may be too weak, too strong, vacuous, or unrelated to the intended task. We therefore evaluate specifications by elaboration, nonvacuity, correct implementation acceptance, wrong implementation rejection, and usefulness for theorem statements or proof attempts.

\subsection{The Theorem and Proof Domain}
\label{sec:theorem_domain}

The Theorem/Proof domain makes correctness claims explicit and tests whether they can be connected to generated implementations and specifications. BRIDGE prompts models to propose properties such as input output relationships, bounds, invariants, monotonicity, preservation, optimality, and termination. These theorem statements are checked in Lean, and in our expanded proof experiments, prover models are asked to complete proofs aligned to the implementation without \texttt{sorry}. This domain is a diagnostic for downstream verification, not a claim that BRIDGE solves proof completion at scale. A theorem statement can elaborate while still being weak, and a completed proof can prove a statement that is too close to the implementation rather than the intended task. We therefore report theorem elaboration, semantic plausibility, nonvacuity, implementation relevance, recurrence across reasoning paths, and proof completion under a fixed prover budget. Appendix~\ref{app:C} gives detailed examples and proof attempt results.

\subsection{Structured Reasoning Hypothesis}
\label{sec:structured_reasoning_hypothesis}

The central hypothesis behind BRIDGE is that verification success depends not only on model capability, but also on the structure of the intermediate reasoning. Direct Lean generation asks the model to recover the algorithm, express it in Lean, satisfy type and termination constraints, infer the intended specification, state useful theorems, and sometimes produce proofs in one step. BRIDGE separates these pressures into linked artifacts. Functional reasoning biases the model toward recursive and compositional definitions that better match Lean. Specification reasoning makes assumptions and edge cases explicit. Theorem and proof reasoning surfaces the formal obligations that a correct implementation should satisfy. This motivates our controls: length controlled prompts test whether verbosity explains the gains, imperative scaffolds test whether any intermediate code reasoning is sufficient, and supervised fine tuning tests whether BRIDGE style traces can be internalized rather than used only at inference time.

\section{Experimental Setup}
\label{sec:experiments}

We evaluate BRIDGE on our 178 problem Lean benchmark and on several public verification settings. The 178 tasks are LeetCode style algorithmic problems, each paired with a natural language statement, a Lean harness, and 100 oracle tests. This gives a common executable criterion across reasoning strategies and lets us isolate the effect of reasoning structure across the Code, Specification, and Theorem/Proof domains. To test transfer, we also evaluate on \textsc{VERINA} with 187 Lean tasks, \textsc{CLEVER} with 161 Lean implementation tasks, and a solver backed \textsc{DafnyBench} pilot. On \textsc{VERINA}, we report pass@1 using the benchmark's \texttt{can\_compile} metric. On \textsc{CLEVER}, we report the executable correctness signal (``OK'') in both pass@1 and larger budget settings. On \textsc{DafnyBench}, we report verification success from the solver. Additional dataset details and benchmark specific protocols are given in Appendix~\ref{app:sample_efficiency}.

Our main Lean study uses five models: \textsc{Claude Sonnet 4}, \textsc{DeepSeek Coder V2}, \textsc{DeepSeek R1}, \textsc{Llama-3.1-70B}, and \textsc{Qwen2.5 Coder}. Unless otherwise stated, models are run with temperature $0.7$ and a maximum output length of $4096$ tokens. Public benchmark and proof experiments additionally use \textsc{Qwen2.5-Coder-32B-Instruct}, \textsc{Gemma4-e4b-it}, \textsc{Goedel-Code-Prover-8B}, and \textsc{DeepSeek-Prover-V2-7B} in the settings reported in Section~\ref{sec:results}. Fine tuning details, including data construction, LoRA settings, and held out evaluation splits, are provided in Appendix~\ref{app:sft_details}.

\textbf{Evaluation protocol.}
For code generation, we evaluate each strategy with $k$ independent samples per task and report pass@$k$, where a task is solved if at least one sample succeeds under the relevant benchmark metric. In refinement settings, each initial sample receives up to three repair rounds. Each repair round appends the Lean error message and asks the model to produce a corrected solution under the same token budget. The ``API'' condition augments the prompt with retrieved Lean v4.21 library signatures and docstrings. Full prompt templates and refinement details are given in Appendix~\ref{app:prompt_system}.

\textbf{Metrics.}
Our main code metric is \emph{Lean executable correctness}: the generated program must elaborate, pass Lean's termination and totality checks, and pass the benchmark tests. For specifications, we separately measure elaboration, nonvacuity, correct implementation acceptance, wrong implementation rejection, and downstream usefulness. For theorem and proof artifacts, we report statement typechecking, semantic plausibility, implementation relevance, and proof completion without \texttt{sorry} where applicable. Detailed proof experiment protocols and examples are given in Appendix~\ref{app:proof_pilot}.

\textbf{Reasoning strategies and controls.}
We compare direct Lean generation with BRIDGE variants that use intermediate reasoning in different paradigms. Functional scaffolds use \textsc{OCaml} or \textsc{Haskell} style reasoning; imperative scaffolds use \textsc{Python}, \textsc{C++}, or \textsc{Java} style reasoning. We also include Double Lean, a length controlled baseline in which the model performs an extra Lean style reasoning stage before producing final Lean code. This control tests whether gains come from the representation itself rather than from longer generations. In the Specification domain, we evaluate Design by Contract, Property Based Testing, Test Driven Development, Defensive Programming, and algorithmic reasoning. In the Theorem/Proof domain, we evaluate generated theorem statements by Lean elaboration, semantic plausibility, recurrence across independent reasoning paths, and bounded proof attempts aligned to the implementation.

\section{Results and Analysis}
\label{sec:results}

\subsection{The Lean Code Domain}
\label{sec:results_code_domain}

We first evaluate BRIDGE in the Code domain, where the goal is to generate executable \textsc{Lean} implementations. Across the BRIDGE suite, functionally aligned intermediate reasoning gives the strongest prompting bias among the strategies we test. Functional scaffolds improve Lean executable correctness by up to $1.5\times$ over direct prompting, and the length controlled Double Lean baseline does not close the gap. This suggests that the gain is not simply due to longer generations or an additional reasoning stage. Rather, functional scaffolds give the model a plan that better matches Lean before it emits code.

\begin{figure}[!ht]
\centering
\begin{subfigure}[t]{0.48\linewidth}
    \centering
    \includegraphics[width=\linewidth]{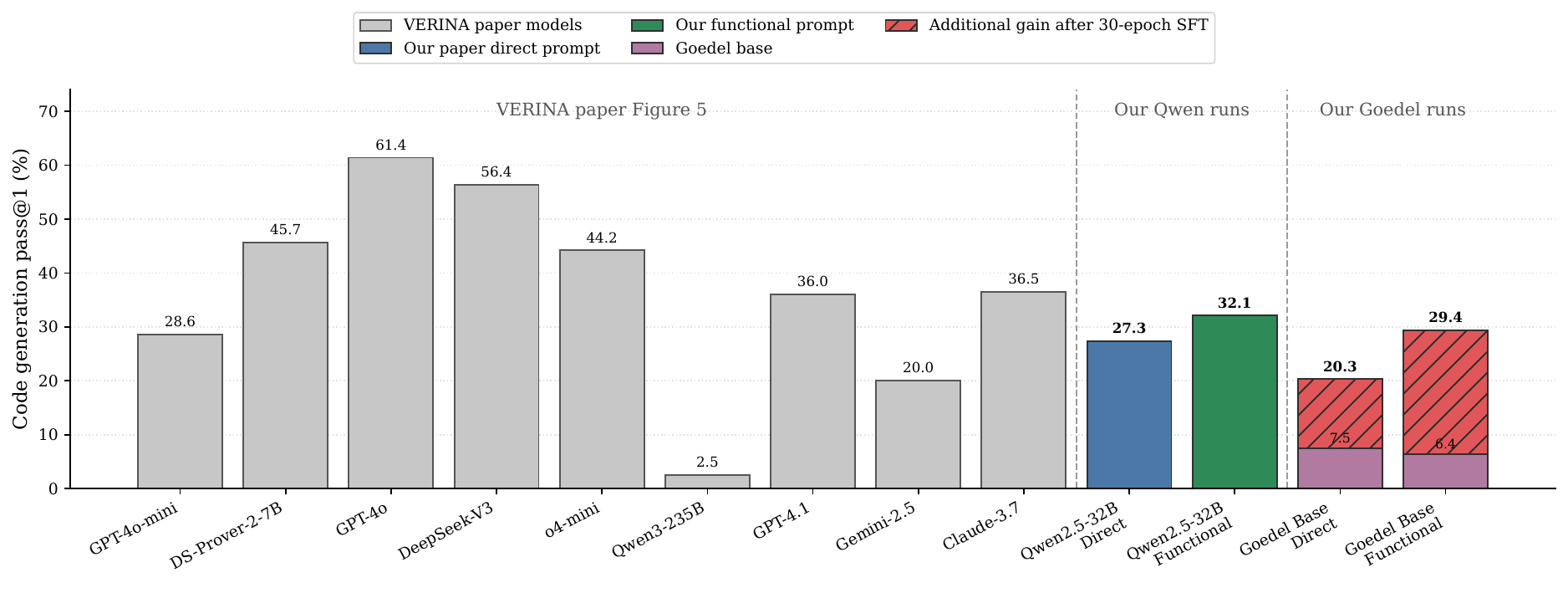}
    \caption{\textsc{VERINA} pass@1.}
\end{subfigure}
\hfill
\begin{subfigure}[t]{0.48\linewidth}
    \centering
    \includegraphics[width=\linewidth]{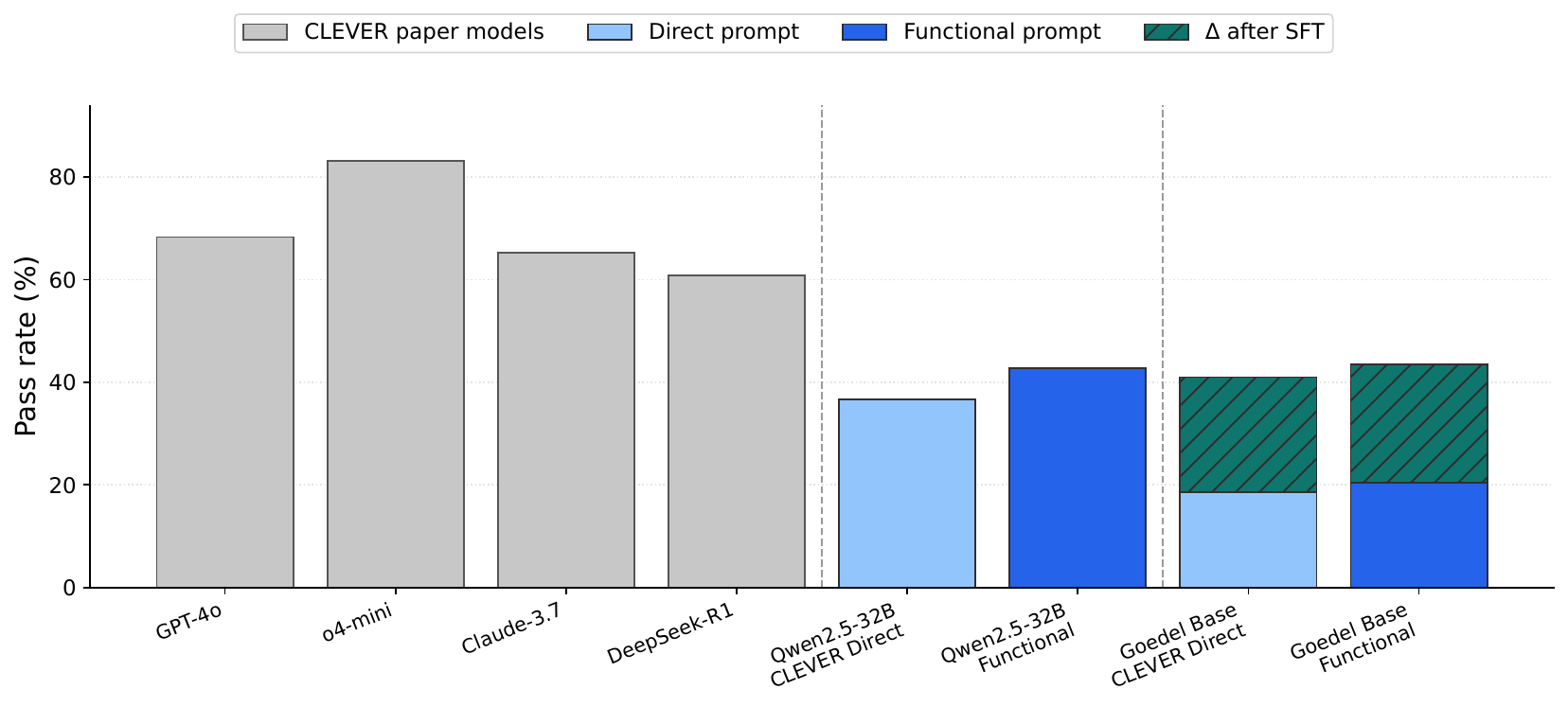}
    \caption{\textsc{CLEVER} executable correctness.}
\end{subfigure}

\vspace{0.35em}

\begin{subfigure}[t]{0.48\linewidth}
    \centering
    \includegraphics[width=\linewidth]{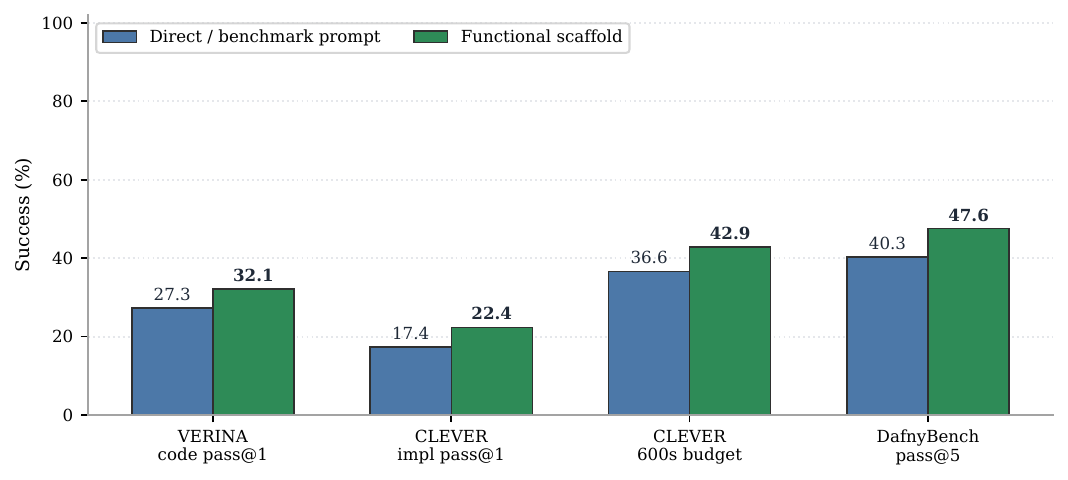}
    \caption{Transfer summary across public verification settings.}
\end{subfigure}
\hfill
\begin{subfigure}[t]{0.48\linewidth}
    \centering
    \includegraphics[width=\linewidth]{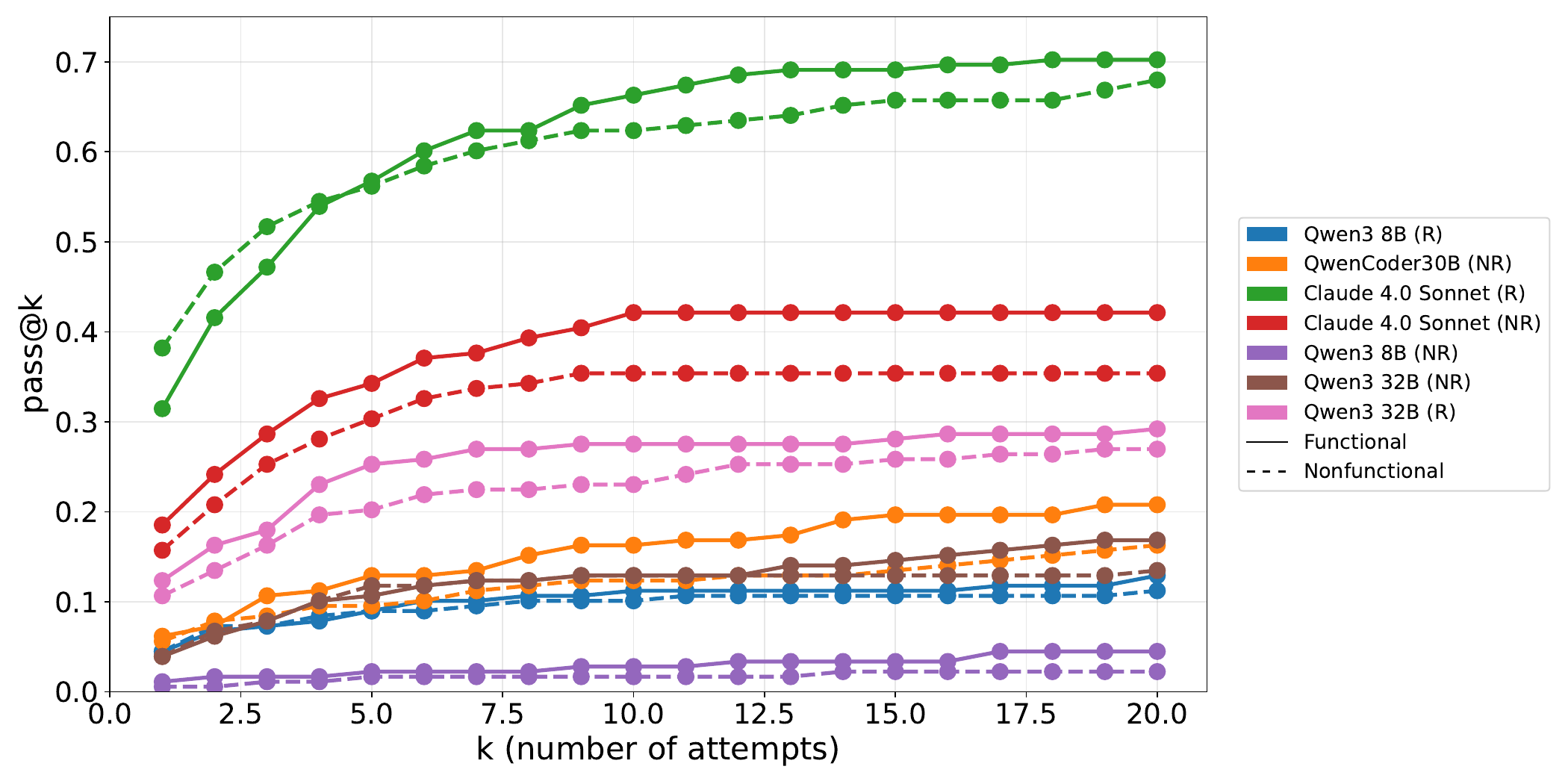}
    \caption{Pass@k scaling on the BRIDGE suite.}
\end{subfigure}

\caption{Code domain results and transfer. Functional scaffolding improves public Lean benchmarks, transfers to verification with solver backends, and reaches comparable BRIDGE suite success with roughly $2\times$ fewer Lean evaluations. Direct uses each benchmark's standard direct prompt; Functional adds the BRIDGE scaffold. Full counts, API ablations, repair round results, and Goedel SFT details are given in Appendix~\ref{app:sample_efficiency} and Appendix~\ref{app:sft_details}.}
\label{fig:code_domain_main}
\end{figure}

The same pattern appears in sample efficiency. As shown in Figure~\ref{fig:code_domain_main}, functional prompting reaches comparable success with roughly $2\times$ fewer Lean evaluations in the pass@k regime. This matters because the main inference cost is not only generating candidates, but also checking them in Lean. Error analysis suggests a mechanism: functional prompting reduces syntax, type, and termination failures, consistent with better alignment to Lean's recursive, typed, and total programming discipline. Detailed ablations with API context and repair rounds are deferred to Appendix~\ref{app:sample_efficiency}.

We next test whether this Code domain effect transfers beyond the BRIDGE suite. On public Lean benchmarks such as \textsc{VERINA} and \textsc{CLEVER}, functional prompting improves pass rates by up to 6.3 percentage points under matched prompt and budget settings. The same scaffold also transfers to code verification with solver backends: on \textsc{DafnyBench}, functional prompting improves verified pass@5 by 7.3 percentage points. These results suggest that BRIDGE is not only a benchmark specific Lean prompt pattern. Functional and specification aware reasoning can help in broader settings where executable code and formal constraints must agree.

\FloatBarrier

Finally, we test whether this structure can be internalized through training. Supervised fine tuning on functional BRIDGE traces improves over matched direct trace fine tuning, most clearly on \textsc{VERINA}, where performance rises from 38/187 to 55/187 (20.3\% to 29.4\%, a 9.1 percentage point gain). Functional SFT also produces shorter and cleaner Lean snippets, with fewer imports, fewer \texttt{sorry}s, and fewer mutation or loop like artifacts. We therefore view BRIDGE as a learnable inductive bias for Lean executable synthesis, not only a one time prompting effect. Full fine tuning settings, counts, and benchmark specific results are reported in Appendix~\ref{app:sft_details}.

\vspace{-0.4cm}
\subsection{The Specification Domain}
\label{sec:results_spec_domain}

\begin{figure}[!ht]
    \centering
    \begin{subfigure}[t]{0.49\linewidth}
        \centering
        \includegraphics[width=\linewidth]{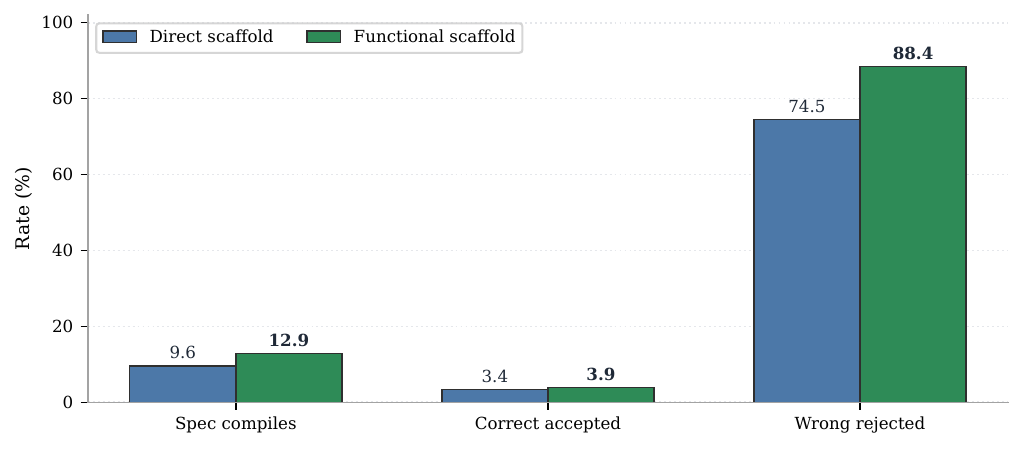}
        \caption{Generated specifications as discriminators of correct and wrong code.}
        \label{fig:spec_discriminative_power_a}
    \end{subfigure}
    \hfill
    \begin{subfigure}[t]{0.49\linewidth}
        \centering
        \includegraphics[width=\linewidth]{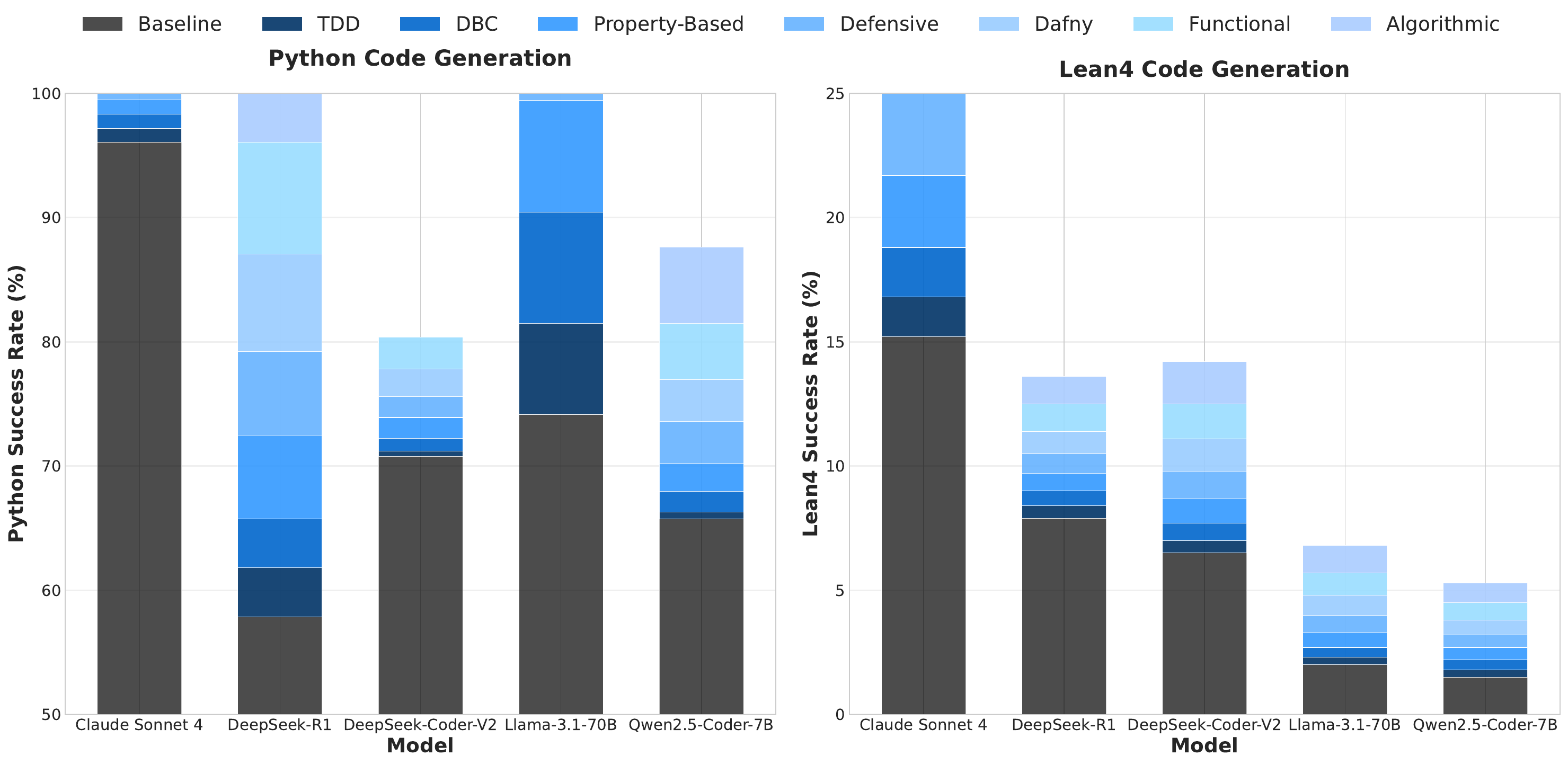}
        \caption{Specification reasoning improves Python and Lean code generation.}
        \label{fig:spec_discriminative_power_b}
    \end{subfigure}
    \caption{Specification domain results. Generated specifications often have useful discriminative power, and specification reasoning also improves code generation. Additional breakdowns are given in Appendix~\ref{app:B}.}
    \label{fig:spec_discriminative_power}
\end{figure}

We next evaluate BRIDGE in the Specification domain, where the goal is not only to generate executable code, but also to make the intended behavior explicit through formal constraints. We study specification reasoning in two ways. First, we ask whether generated specifications have \emph{semantic bite}: can they help distinguish correct implementations from reasonable wrong ones? Second, we ask whether reasoning through specifications improves code generation itself. Figure~\ref{fig:spec_discriminative_power}(a) shows that generated specifications are often useful as discriminators, indicating that they are not merely syntactically valid or vacuous. This matters because a specification that typechecks but accepts both correct and incorrect code contributes little to verification.

Figure~\ref{fig:spec_discriminative_power}(b) shows the second effect: specification reasoning also improves code generation. In conventional Python generation, specification reasoning yields gains of up to 17.5 percentage points, and it also improves Lean code generation, although verified Lean success remains the harder bottleneck. This separation is useful. Specification reasoning helps both as a planning tool for code generation and as a source of downstream formal constraints, even when full end to end verification remains difficult. Additional specification styles, detailed quantitative breakdowns, and cross language examples are given in Appendix~\ref{app:B}.

\FloatBarrier

\begin{figure}[!ht]
\centering
\begin{subfigure}[!t]{0.48\linewidth}
    \centering
    \includegraphics[width=\linewidth]{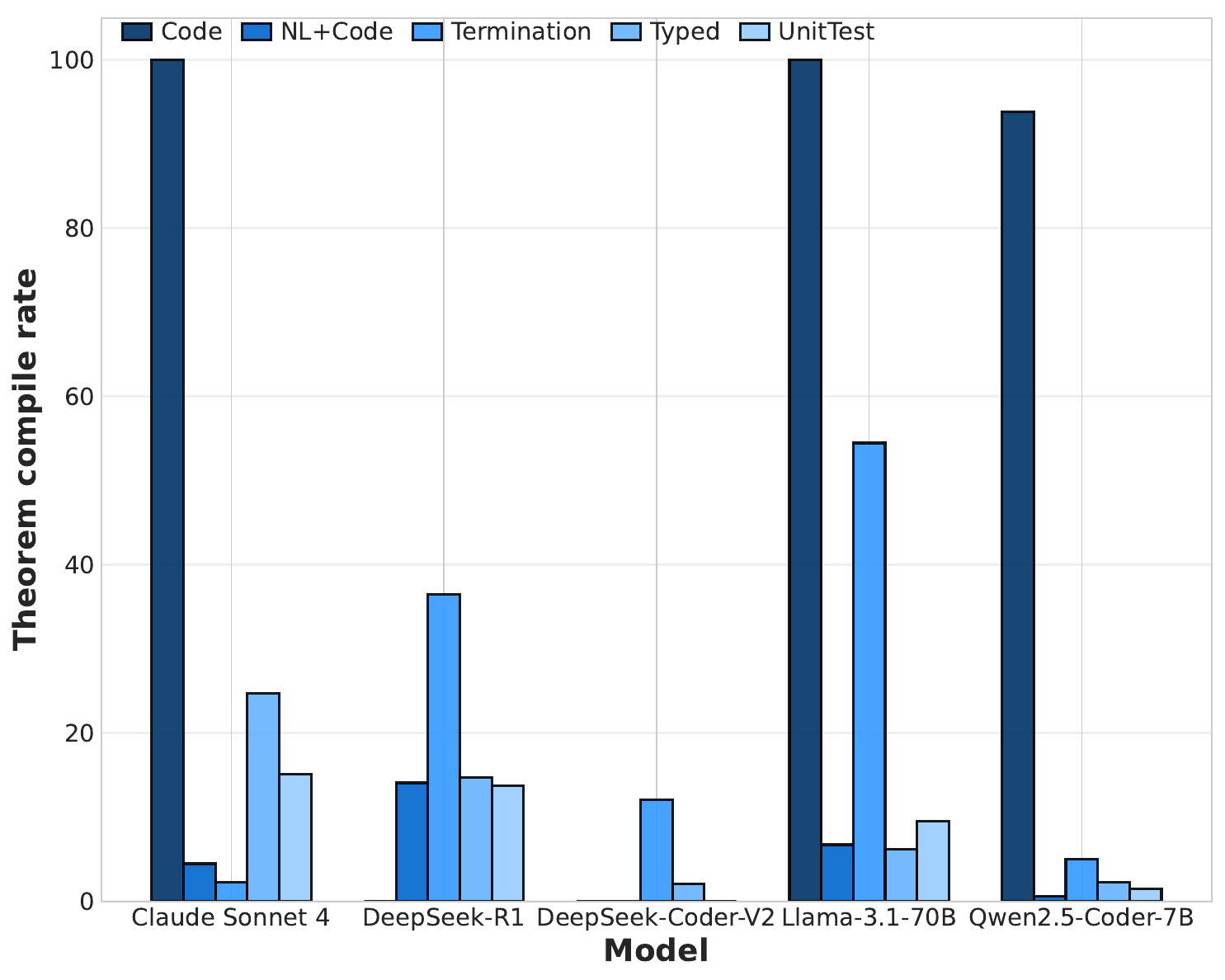}
    \caption{Theorem statement elaboration.}
    \label{fig:theorem_proof_diagnostics_a}
\end{subfigure}
\hfill
\begin{subfigure}[!t]{0.48\linewidth}
    \centering
    \includegraphics[width=\linewidth]{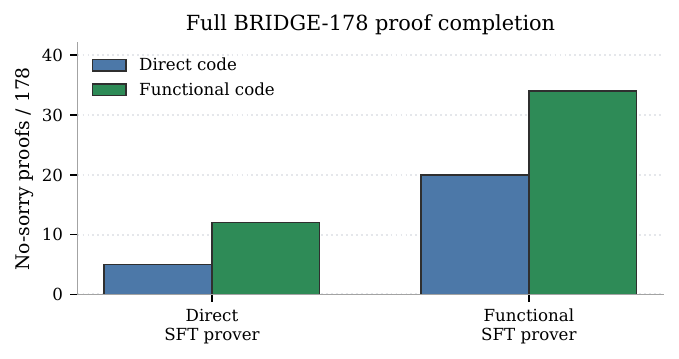}
    \caption{Implementation aligned proof attempts across provers.}
    \label{fig:theorem_proof_diagnostics_b}
\end{subfigure}
\begin{subfigure}[!t]{0.48\linewidth}
    \centering
    \includegraphics[width=\linewidth]{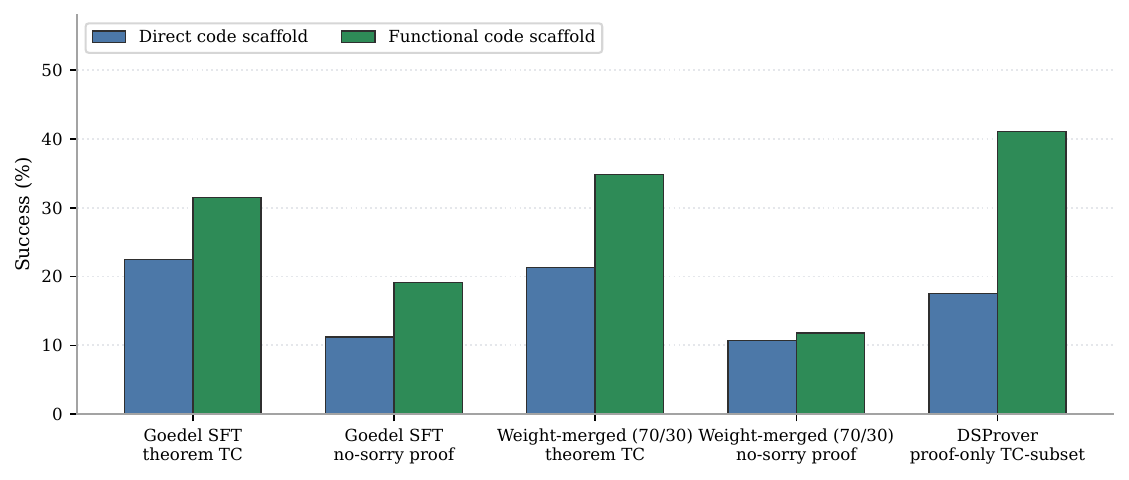}
    \caption{Proof only DSProver check.}
    \label{fig:theorem_proof_diagnostics_c}
\end{subfigure}
\hfill
\begin{subfigure}[!t]{0.48\linewidth}
    \centering
    \includegraphics[width=\linewidth]{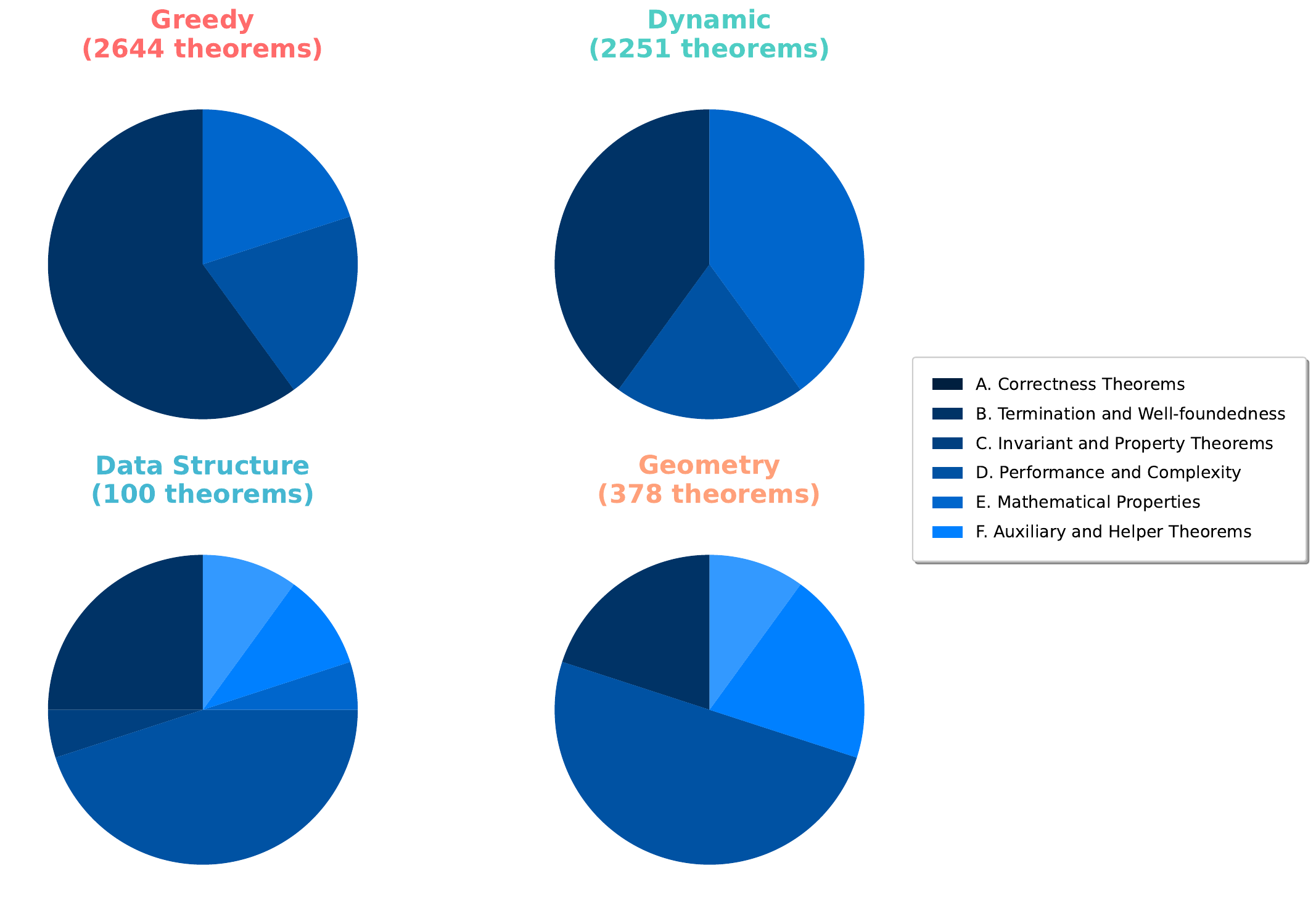}
    \caption{Distribution of generated theorem categories.}
    \label{fig:theorem_proof_diagnostics_d}
\end{subfigure}

\caption{Theorem/proof diagnostics. Functional scaffolds improve theorem elaboration, proof completion, and proof only success; the category plot summarizes generated theorem families.}
\label{fig:theorem_proof_diagnostics}
\end{figure}
\FloatBarrier
\vspace{-0.4cm}
\subsection{The Theorem and Proof Domain}
\label{sec:results_theorem_proof_domain}

We finally evaluate theorem statements and proof attempts as diagnostics for downstream verification. Functional scaffolds make the generated Lean code easier to connect to formal claims: they yield more Lean elaborating theorem statements aligned to the implementation than direct prompting, and they also produce shorter implementations that are easier for proof models to inspect. In the full 178 task proof experiment, functional code gives more completed proofs without \texttt{sorry} under the same prover budget, including 34/178 versus 20/178 with the functional SFT prover. This suggests that the benefit of BRIDGE is not limited to executable code. Better code structure also improves the formal artifacts that can be attached to that code. We further isolate proof search from theorem statement generation using \textsc{DeepSeek-Prover-V2-7B} as a proof only model on statements that already elaborate.  This check shows the same direction: DSProver completes 23/56 functional proofs versus 7/40 direct proofs without \texttt{sorry}. Thus functional scaffolding appears to help both stages of the theorem/proof pipeline: producing statements that Lean accepts, and making already accepted statements easier for a prover to discharge. These results do not imply full semantic verification, since claims aligned to the implementation can still be weaker than the original task specification. Instead, they show that better code structure improves the quality and provability of the formal artifacts attached to it. Semantic specification variants, additional prover baselines, and qualitative examples are given in Appendix~\ref{app:proof_pilot}.
\vspace{-0.3cm}
\section{Limitations}
\label{sec:limitations}

BRIDGE studies the \emph{front end} of verified synthesis, not full semantic verification. Lean executable correctness checks elaboration, termination and totality acceptance, and benchmark tests, but these checks can still miss the intended specification. Generated specifications and theorem statements may also typecheck while being weak, vacuous, or overly shaped by the implementation. We therefore report diagnostics such as nonvacuity, implementation relevance, wrong implementation rejection, proof attempts, and manual review, but these do not eliminate false confidence. Our evaluation includes the BRIDGE suite, public Lean benchmarks, and a solver backed \textsc{DafnyBench} pilot, but the tasks are still mostly local and algorithmic. We have not yet tested BRIDGE on large formal libraries, rich stateful or effectful programs, long horizon proofs, or multi module verification. Results also vary across models, prompts, and search budgets, so BRIDGE should be viewed as a useful inductive bias rather than a universal improvement.
\vspace{-0.4cm}
\section{Conclusion and Future Work}
\label{sec:conclusion}

We introduced BRIDGE, a framework that reframes verifiable coding as the problem of maintaining semantic consistency across interdependent code, specification, theorem, and proof artifacts. Rather than asking a model to jump directly from natural language to a final proof, BRIDGE decomposes the process into structured domains whose artifacts can constrain and check one another. The main lesson from our experiments is that \emph{reasoning representation matters}: functionally aligned intermediate scaffolds improve Lean executable correctness relative to direct prompting and length controlled baselines, reduce syntax, type, and termination failures, and reach comparable success with fewer Lean evaluations.

The specification and theorem/proof results show that this effect is not limited to code generation. Specification reasoning makes assumptions, edge cases, and behavioral constraints explicit, improving both conventional code generation and the usefulness of generated formal constraints. Theorem and proof artifacts provide an additional diagnostic: functional code is easier to attach Lean elaborating theorem statements to, and easier for proof models to discharge under fixed budgets. The transfer and fine tuning results further support this view. BRIDGE style functional prompting improves public Lean benchmarks such as \textsc{VERINA} and \textsc{CLEVER}, transfers to verification with solver backends through \textsc{DafnyBench}, and can be internalized through supervised fine tuning on structured traces. This indicates that BRIDGE is not merely a prompt trick for one benchmark, but a learnable inductive bias for connecting programs, specifications, theorem statements, and proof attempts.

The natural next step is to close the loop between BRIDGE artifacts and stronger proof search. Future systems should combine structured artifact generation with tool using theorem provers, formal library retrieval, verifier guided repair, and post training methods such as reinforcement learning from verifiable rewards. This direction would move LLMs toward producing checkable, progressively verifiable artifacts rather than merely plausible code, making formal verification a more integrated part of everyday code generation.

\vspace{-0.4cm}
\section*{Broader Impact}
This work may reduce the cost of producing formal programming artifacts and make proof assistant workflows more accessible. Even when full proofs are not yet available, better code, specification, theorem, and proof drafts can help experts prototype, audit, and refine verification tasks more quickly. 

\bibliographystyle{plainnat}

\bibliography{references}

\appendix

\section{Additional Code Domain and Public Benchmark Details}
\label{app:sample_efficiency}

This section gives the detailed Code domain evidence behind the main results. The main paper reports the headline effects: functional scaffolds improve Lean executable correctness, reach comparable success with fewer Lean evaluations, and transfer to public Lean and solver backed verification settings. Here we provide the detailed ablations, public benchmark counts, and generation length analysis.

\subsection{BRIDGE Suite Ablations}

\begin{figure}[!ht]
\centering
\includegraphics[width=0.78\linewidth]{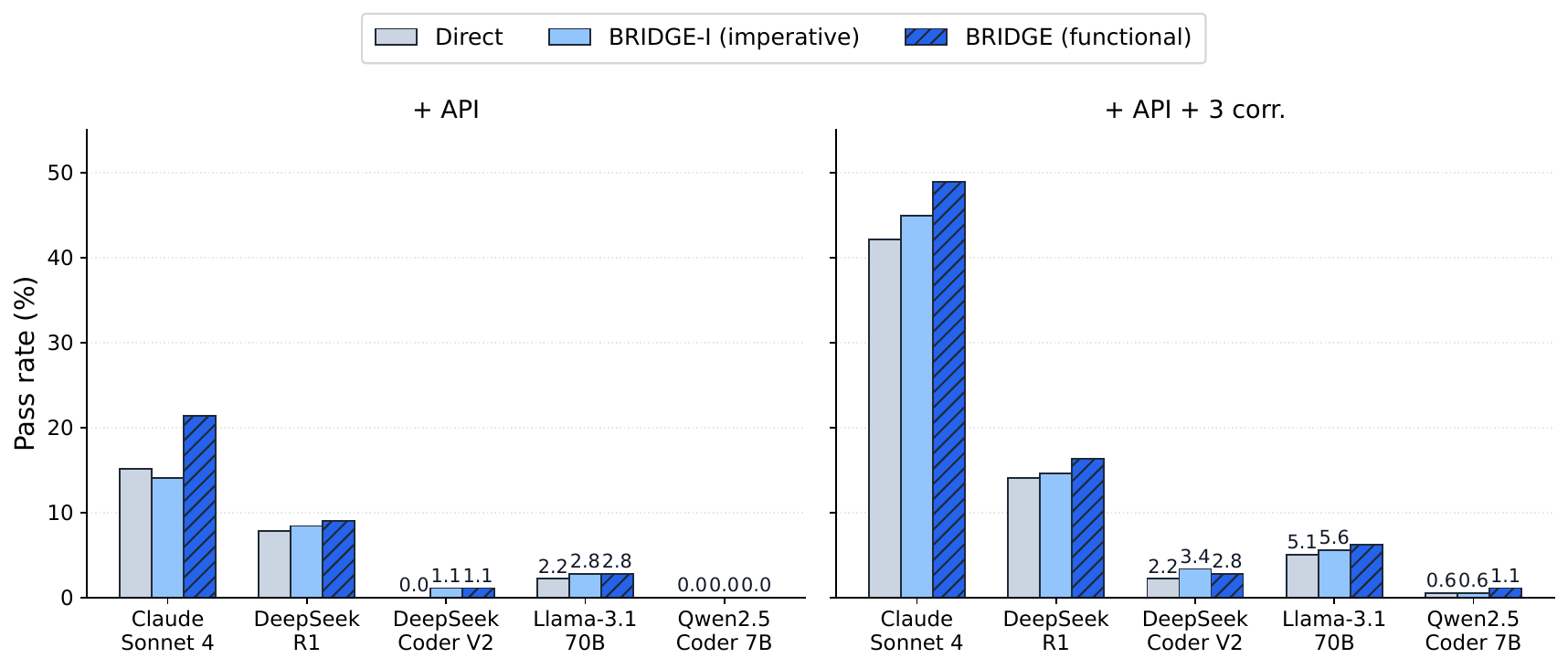}
\caption{Detailed BRIDGE suite ablations with API context and repair rounds. Functional scaffolding remains the strongest structured prompting condition, while API context and repair rounds narrow but do not remove the representation effect.}
\label{fig:appendix_ablation_breakdown}
\end{figure}

Figure~\ref{fig:appendix_ablation_breakdown} gives the detailed ablation breakdown on the 178 problem BRIDGE suite. The main paper keeps the pass@k and transfer plots as headline figures, while this appendix figure shows the API context and repair round variants. The length controlled Double Lean control is important because it has comparable generation length to functional scaffolding but does not close the performance gap. This supports the interpretation that the gain comes from the reasoning representation rather than verbosity alone.

\FloatBarrier

\subsection{Public Benchmark Transfer Counts}

Table~\ref{tab:public_transfer_details} gives the full counts behind the transfer results summarized in the main text. Functional scaffolding improves most matched prompt and budget settings, including Qwen on \textsc{VERINA} and \textsc{CLEVER}, larger budget \textsc{CLEVER}, and solver backed \textsc{DafnyBench}. The main exception is base Goedel on \textsc{VERINA}, where functional prompting slightly underperforms direct prompting. We treat this as evidence that the scaffold is a useful inductive bias, not a universal prompt improvement independent of model and calibration.

\begin{table}[!ht]
\centering
\caption{Direct versus functional transfer results on public verification benchmarks. Functional scaffolding improves most matched settings, but gains remain model and budget dependent.}
\label{tab:public_transfer_details}
\small
\setlength{\tabcolsep}{4pt}
\begin{tabular}{llccc}
\toprule
\textbf{Benchmark} & \textbf{Model and setting} & \textbf{Direct} & \textbf{Functional} & \textbf{$\Delta$} \\
\midrule
\textsc{VERINA} & Qwen2.5 Coder pass@1 & 51/187 & 60/187 & +4.8 pp \\
\textsc{VERINA} & Base Goedel pass@1 & 14/187 & 12/187 & -1.1 pp \\
\textsc{CLEVER} & Qwen pass@1 & 28/161 & 36/161 & +5.0 pp \\
\textsc{CLEVER} & Qwen larger budget & 59/161 & 69/161 & +6.3 pp \\
\textsc{CLEVER} & Base Goedel larger budget & 66/161 & 70/161 & +2.5 pp \\
\textsc{CLEVER} & Goedel SFT e30 larger budget & 30/161 & 33/161 & +1.9 pp \\
\textsc{DafnyBench} & Solver backed pass@5 & 315/782 & 372/782 & +7.3 pp \\
\bottomrule
\end{tabular}
\end{table}

\subsection{Generation Length and Sample Efficiency}

Functional scaffolds are longer than direct generations, but length alone does not explain the gains. Table~\ref{tab:generation_length} reports average word and token counts by strategy. Double Lean is close to the functional methods in length, yet performs worse in the main experiments. This supports the main claim: the useful signal is not simply more text, but a scaffold whose structure better matches Lean.

\begin{table}[!ht]
\centering
\caption{Generation lengths by reasoning strategy. Lengths are reported as words and tokens, averaged over all attempts, successful attempts, and failed attempts.}
\label{tab:generation_length}
\footnotesize
\renewcommand{\arraystretch}{1.15}
\setlength{\tabcolsep}{2pt}
\begin{tabular}{lccc}
\toprule
\textbf{Reasoning Strategy} &
\textbf{Average} &
\textbf{Successful} &
\textbf{Failed} \\
\midrule
Direct & 195 / 270 & 141 / 196 & 200 / 278 \\
Double Lean & 402 / 555 & 347 / 480 & 412 / 570 \\
Python & 382 / 526 & 324 / 446 & 391 / 540 \\
Haskell & 387 / 534 & 334 / 460 & 401 / 555 \\
C++ & 390 / 538 & 341 / 469 & 402 / 558 \\
OCaml & 408 / 563 & 343 / 473 & 421 / 581 \\
\bottomrule
\end{tabular}
\end{table}

The practical implication is that functional scaffolds can trade a larger per sample generation for fewer Lean evaluations. Since Lean checking and iterative repair dominate the workflow cost, sample efficiency is a meaningful advantage even when individual generations are longer.

\section{Specification and Representation Details}
\label{app:B}

Here we focus on how BRIDGE uses specification reasoning to make task intent explicit and to test whether generated constraints are meaningful. The goal is not to treat a generated specification as a final certificate of correctness. Instead, specifications act as intermediate formal artifacts that constrain implementations, expose hidden assumptions, and provide targets for theorem statements and proof attempts.

\subsection{Programming Paradigms and Specification Structure}

In \textsc{Lean}, programs are mathematical objects inside dependent type theory, so the structure of the implementation strongly affects the structure of the specification. A functional implementation usually exposes recursive structure, base cases, and compositional definitions directly. This makes it easier to state properties by induction or by rewriting. An imperative implementation may be algorithmically natural, but loops, mutation, and index updates often require additional invariants before the intended behavior can be expressed cleanly.

This distinction matters for specifications. A specification for a recursive function can often mirror the recursive decomposition of the code. A specification for an imperative loop must usually describe the relationship between initial state, loop state, and final state. BRIDGE uses intermediate scaffolds to bias the model toward artifacts that are easier to connect: functional scaffolds expose recursion and invariants early, while specification scaffolds expose assumptions and expected behavior before or after implementation.

\begin{table}[!ht]
\centering
\caption{Specification reasoning styles used in BRIDGE. Each style exposes a different view of task intent and a different failure mode.}
\label{tab:spec_reasoning_styles}
\small
\setlength{\tabcolsep}{4pt}
\begin{tabular}{lll}
\toprule
\textbf{Style} & \textbf{Main artifact} & \textbf{Primary risk} \\
\midrule
Design by Contract & Preconditions and postconditions & Missing relational properties \\
Property Reasoning & Bounds, monotonicity, preservation & Weak or irrelevant properties \\
Test Driven Reasoning & Edge case constraints & Overfitting to examples \\
Defensive Reasoning & Input assumptions and safe behavior & Excessive guards \\
Algorithmic Reasoning & Decomposition and invariants & Informal constraints only \\
\bottomrule
\end{tabular}
\end{table}

\subsection{Specification Reasoning Styles}

\paragraph{Design by Contract.}
Design by Contract asks the model to identify preconditions, postconditions, and invariants before writing or checking the implementation. In Lean, these usually become hypotheses, predicates over inputs and outputs, or auxiliary invariant definitions. This style is useful for making hidden assumptions explicit. Its main limitation is that local contracts may miss global or relational properties, such as optimality, permutation equivalence, or behavior across multiple calls.

\paragraph{Property reasoning.}
Property reasoning asks the model to identify universal behavior such as bounds, monotonicity, preservation, symmetry, idempotence, or algebraic laws. These properties often become useful theorem candidates. However, they can also be too weak. For example, a theorem that only states that an output is nonnegative may elaborate and even be provable, while saying very little about the original task.

\paragraph{Test driven reasoning.}
Test driven reasoning uses examples and edge cases as a way to infer intended behavior. This is useful for tasks with many boundary cases, such as empty inputs, singleton lists, duplicate elements, or degenerate numerical ranges. The limitation is that tests can suggest an incomplete specification. BRIDGE therefore treats test driven reasoning as a source of candidate constraints, not as a replacement for formal semantics.

\paragraph{Defensive reasoning.}
Defensive reasoning asks the model to surface invalid inputs, hidden assumptions, and safe fallback behavior. This can prevent vacuous specifications that silently assume away difficult cases. The risk is that excessive guards can obscure the core mathematical property or make the specification too narrow.

\paragraph{Algorithmic reasoning.}
Algorithmic reasoning asks the model to describe the intended computational structure before formalizing the specification. This is useful when the behavior is easiest to characterize through an invariant or recurrence. For example, a dynamic program may require a specification over subproblem values rather than only a final input output relation.

\subsection{Specification Evaluation Signals}

A generated specification can elaborate while still being semantically unhelpful. BRIDGE therefore evaluates specifications using multiple signals rather than a single binary label.

\begin{table}[!ht]
\centering
\caption{Specification evaluation signals. BRIDGE uses these diagnostics to distinguish syntactic validity from semantic usefulness.}
\label{tab:spec_eval_signals}
\small
\setlength{\tabcolsep}{4pt}
\begin{tabular}{ll}
\toprule
\textbf{Signal} & \textbf{Purpose} \\
\midrule
Elaboration & Checks whether the specification is valid Lean. \\
Nonvacuity & Checks whether the specification constrains behavior. \\
Correct implementation acceptance & Checks whether the intended implementation satisfies the spec. \\
Wrong implementation rejection & Checks whether plausible incorrect code is ruled out. \\
Downstream usefulness & Checks whether the spec supports theorem or proof attempts. \\
\bottomrule
\end{tabular}
\end{table}

The discriminator setting is particularly important. A useful specification should accept a correct implementation while rejecting plausible wrong implementations. This is stricter than elaboration alone. In our specification discriminator experiment, functional scaffolding improves specification compilation and wrong implementation rejection. Functional specifications reject 61/69 wrong variants, or 88.4\%, compared with 38/51, or 74.5\%, for direct specifications. Correct implementation acceptance remains low for both conditions, 7/178 for functional and 6/178 for direct, which is a useful negative result: broad, faithful semantic specification generation remains substantially harder than executable code generation.

\subsection{Specification Reasoning as a Code Generation Tool}

Specification reasoning also improves code generation itself. In conventional Python generation, specification guided prompting improves pass rates by up to 17.5 percentage points. This suggests that specifications help as a planning tool even when the final output is ordinary executable code. The same effect appears in Lean, but verified Lean success remains the harder bottleneck because the generated code must satisfy typing, totality, termination, and tests.

This separation is important. Specification reasoning has two roles in BRIDGE. First, it helps models plan better implementations by making edge cases and constraints explicit. Second, it produces formal artifacts that can be checked against implementations and later theorem statements. The first role improves code generation; the second supports downstream verification.

\subsection{Embedding Analysis of Generated Code}
\label{app:embedding_analysis}

\begin{figure}[!ht]
    \centering
    \includegraphics[width=0.58\linewidth]{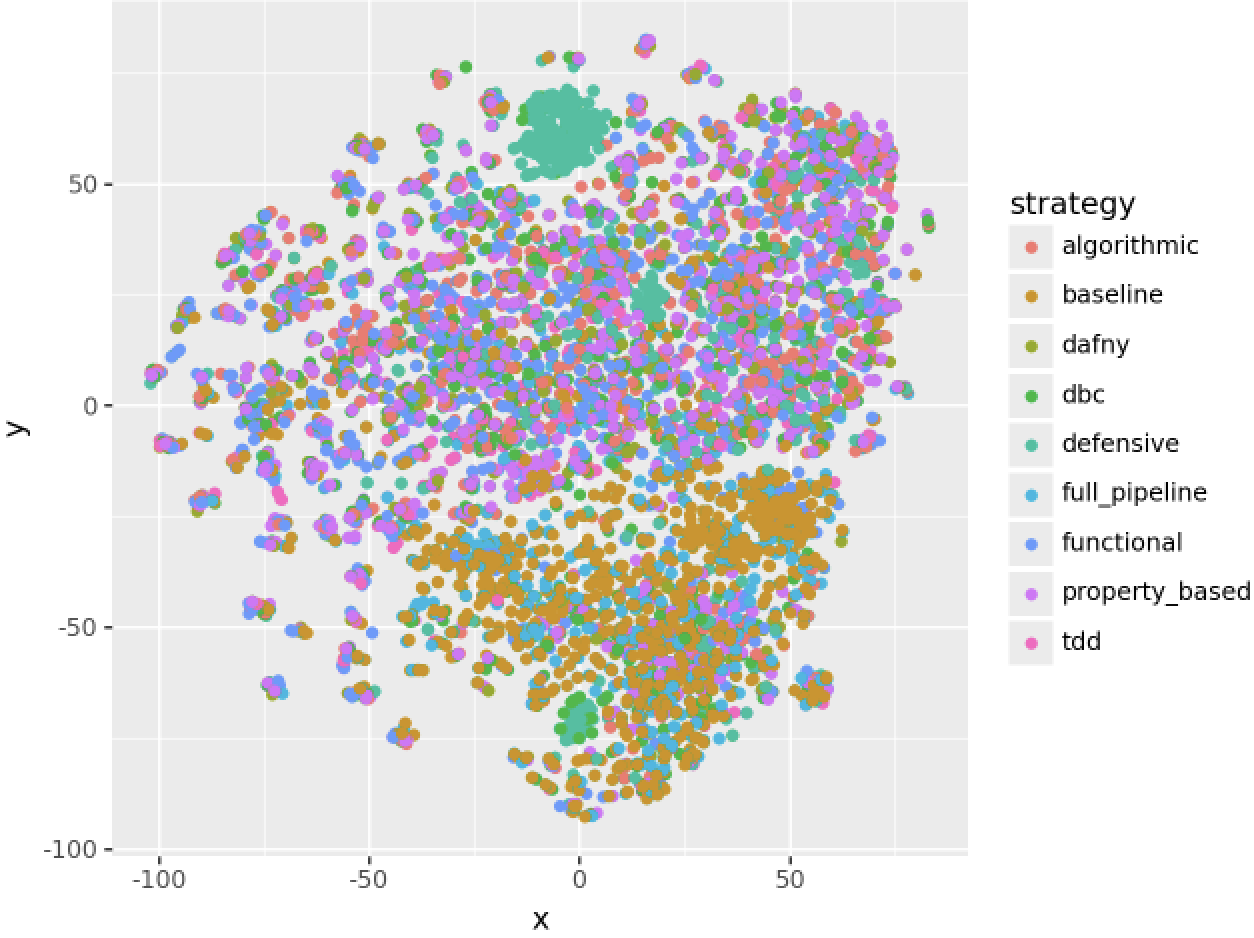}
    \caption{Embedding analysis of generated Python implementations. Structured reasoning strategies cluster more tightly, while direct baseline solutions are more dispersed.}
    \label{fig:embedding_analysis}
\end{figure}

To test whether structured reasoning changes the generated implementation distribution, we embed generated Python implementations using Qwen2.5 Coder embeddings and visualize them with t-SNE. Figure~\ref{fig:embedding_analysis} shows that structured strategies cluster together, including Design by Contract, defensive reasoning, functional reasoning, and property based reasoning. Direct baseline generations are more dispersed.

\FloatBarrier

This supports the interpretation that BRIDGE changes the implementation distribution, not only the surrounding explanation. Different scaffolds lead the model toward different implementation families, which then affects whether specifications and theorem statements can attach cleanly.

\section{Theorem and Proof Details}
\label{app:proof_pilot}
\label{app:C}

Here we focus on theorem statements and proof attempts as diagnostics for downstream verification. The key distinction is between three levels of success: stating a claim that elaborates in Lean, proving that claim without \texttt{sorry}, and proving a faithful semantic specification of the original task. BRIDGE improves the first two signals, while the third remains the harder open problem.

\subsection{Multi Pathway Theorem Discovery}

Theorem generation is difficult because the model must decide which mathematical facts make the implementation correct. BRIDGE uses multiple pathways to reduce the chance that a single prompt produces a shallow or implementation shaped theorem.

\begin{table}[!ht]
\centering
\caption{Theorem discovery pathways used in BRIDGE. Each pathway surfaces a different kind of formal obligation.}
\label{tab:theorem_pathways}
\small
\setlength{\tabcolsep}{4pt}
\begin{tabular}{ll}
\toprule
\textbf{Pathway} & \textbf{Typical theorem content} \\
\midrule
Natural language & Functional correctness and input output relationships \\
Unit tests & Generalized edge case properties and invariants \\
Code analysis & Implementation aligned claims and structural invariants \\
Type guided reasoning & Properties suggested by Lean types and dependencies \\
Termination reasoning & Measures, recursion, and complexity bounds \\
\bottomrule
\end{tabular}
\end{table}

The pathways are complementary. Natural language prompts tend to produce high level correctness claims. Unit test prompts often identify boundary behavior. Code analysis prompts generate aligned to the implementation statements that are easier to prove, but may be weaker than the intended task. Type guided prompts can exploit Lean's dependent structure, while termination prompts expose the measures required for recursive definitions.

\subsection{Full Implementation Aligned Proof Pilot}

To test whether functional scaffolds make downstream verification artifacts easier to build, we run a full 178 task aligned to the implementation proof experiment. For each task, we select matched direct and functional Lean candidates. A prover then generates a specification and theorem tied to the candidate implementation and attempts to complete the proof. We score statement typechecking, strict nonvacuity, proof typechecking, proof completion without \texttt{sorry}, and implementation length.

\begin{table}[!ht]
\centering
\caption{BRIDGE suite theorem, specification, and proof experiment on all 178 tasks. Implementation aligned specifications condition on the candidate implementation. Functional code is shorter and yields more theorem statements and completed proofs across multiple prover checkpoints.}
\label{tab:proof_pilot}
\small
\setlength{\tabcolsep}{3pt}
\resizebox{\linewidth}{!}{%
\begin{tabular}{llccccc}
\toprule
\textbf{Prover} & \textbf{Code Style} & \textbf{Stmt TC} & \textbf{Strict Nonvac.} & \textbf{Proof TC} & \textbf{No \texttt{sorry} Proof} & \textbf{Avg Code Chars} \\
\midrule
Direct SFT & Direct & 11/178 & 2/178 & 5/178 & 5/178 & 532.7 \\
Direct SFT & Functional & 25/178 & 1/178 & 12/178 & 12/178 & 343.5 \\
Functional SFT & Direct & 40/178 & 2/178 & 20/178 & 20/178 & 532.7 \\
Functional SFT & Functional & 56/178 & 4/178 & 34/178 & 34/178 & 343.5 \\
Weight merged 70/30 & Direct & 38/178 & 1/178 & 19/178 & 19/178 & 532.7 \\
Weight merged 70/30 & Functional & 62/178 & 2/178 & 21/178 & 21/178 & 343.5 \\
\bottomrule
\end{tabular}
}
\end{table}

\begin{figure}[!ht]
\centering
\begin{subfigure}[t]{0.48\linewidth}
    \centering
    \includegraphics[width=\linewidth]{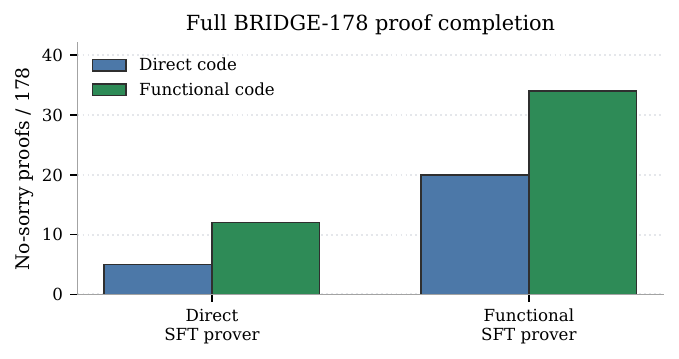}
    \caption{Implementation aligned proof pilot across provers.}
    \label{fig:proof_rates_a}
\end{subfigure}
\hfill
\begin{subfigure}[t]{0.48\linewidth}
    \centering
    \includegraphics[width=\linewidth]{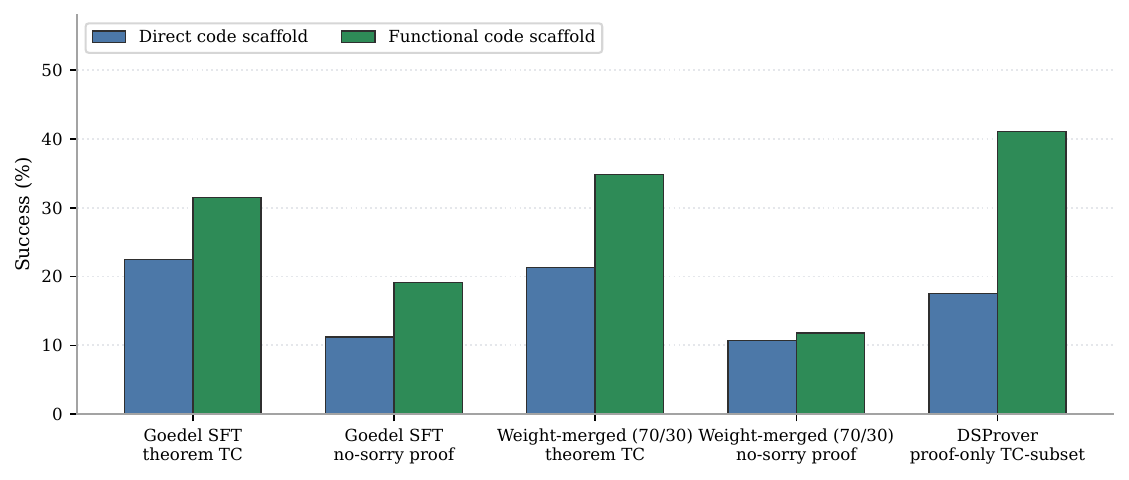}
    \caption{Proof only DSProver success rates.}
    \label{fig:proof_rates_b}
\end{subfigure}
\caption{Proof diagnostics on the BRIDGE suite. Functional code yields more elaborating theorem statements and more completed proofs under matched prover budgets.}
\label{fig:proof_rates}
\end{figure}
\FloatBarrier

The aligned to the implementation result is useful as a proof effort diagnostic. It asks whether a generated implementation gives the prover enough structure to state and discharge checkable claims. Functional code consistently gives the prover shorter implementations and more successful statement/proof attempts. However, aligned to the implementation specifications can still be weaker than the natural language task, so these results should not be read as full semantic verification.

\subsection{Proof Only DSProver Check}

We also run \textsc{DeepSeek-Prover-V2-7B} as a proof only model on aligned to the implementation theorem statements that already elaborate. This isolates proof search from theorem statement generation. The direct code subset contains 40 typechecked statements, and the functional code subset contains 56. DSProver completes 7/40 direct proofs and 23/56 functional proofs without \texttt{sorry}.

\begin{table}[!ht]
\centering
\caption{Proof only DSProver check on aligned to the implementation theorem statements that already elaborate. Functional code gives the proof model a larger and easier to prove theorem set.}
\label{tab:dsprover_proof_only}
\small
\setlength{\tabcolsep}{5pt}
\begin{tabular}{lccc}
\toprule
\textbf{Code Style} & \textbf{Input Statements} & \textbf{No \texttt{sorry} Proofs} & \textbf{Rate} \\
\midrule
Direct & 40 & 7 & 17.5\% \\
Functional & 56 & 23 & 41.1\% \\
\bottomrule
\end{tabular}
\end{table}

This proof only setting shows that functional scaffolding helps even after theorem statement generation is fixed. The proof model receives statements that already elaborate, and functional code still yields a higher proof completion rate. This suggests that the structure of the implementation itself matters for downstream proof search.

A raw \textsc{VERINA} proof only DSProver adapter run was not used as a headline result. Although it completed 187 proof tasks, outputs often violated the expected structured format or contained placeholders, yielding 0/187 accepted proofs. We therefore report DSProver only in the setting where the Lean splice and scoring format are explicit.

\subsection{Weight Merging Versus SFT}

We also test a simple alternative to supervised fine tuning: linear weight merging. Specifically, we merge \textsc{Goedel-Code-Prover-8B}, which is closer to the code generation setting, with \textsc{Goedel-Prover-V2-8B}, which has stronger mathematical proof priors, using a 70/30 mixture. The motivation is that Lean sits between programming and mathematics. A model must produce syntactically valid code, but it must also state mathematical claims, use proof idioms, and manipulate propositions in a theorem prover. Merging with a math oriented prover may therefore improve the theorem statement side of the pipeline without requiring new training.

The results support this interpretation, but also show its limits. The merged prover improves theorem statement typechecking for functional code, reaching 62/178 statement typechecks compared with 38/178 for direct code under the same merged prover. This is the strongest statement typechecking result among the prover variants. However, it does not match the functional SFT prover on completed proofs: the merged prover completes 21/178 proofs without \texttt{sorry} for functional code, whereas the functional SFT prover completes 34/178. Thus, merging with a math model appears especially useful for producing Lean elaborating theorem statements, while BRIDGE style SFT is more effective for turning those statements into completed proofs aligned to the implementation.

This suggests a useful division of labor. Weight merging is a cheap way to inject mathematical proof priors into a code prover, and it may be especially helpful for Lean because Lean artifacts require both programming structure and theorem proving structure. But SFT on functional BRIDGE traces provides a more targeted alignment signal: it teaches the model not only to speak Lean mathematics, but to connect functional implementations, specifications, theorem statements, and proof attempts in the format required by the task.

\subsection{Completed Micro Proofs}

The following examples illustrate the artifact stack used in the proof experiment. These are aligned to the implementation proofs: they show that the generated implementation satisfies the generated contract. They are useful diagnostics for code, specification, and proof alignment, but they should not be interpreted as deep semantic proofs of the original tasks.

\begin{lstlisting}[style=verbatimish, language=Lean, caption={A compact aligned to the implementation proof.}]
def minPushes (w : String) : Nat :=
  (List.range w.length).foldl
    (fun acc i => acc + i / 8 + 1) 0

def ValidWord (w : String) : Prop :=
  1 <= w.length ∧ w.length <= 26 ∧ w.data.Nodup

def MinPushesSpec (w : String) (out : Nat) : Prop :=
  ValidWord w →
  out =
    (List.range w.length).foldl
      (fun acc i => acc + i / 8 + 1) 0

theorem minPushes_satisfies_spec (w : String) :
  MinPushesSpec w (minPushes w) := by
  intro _
  unfold MinPushesSpec minPushes
  rfl
\end{lstlisting}

The next example gives a lightweight invariant theorem. It is not a full task level semantic specification, but it is nonvacuous and implementation relevant.

\begin{lstlisting}[style=verbatimish, language=Lean, caption={A simple invariant proof attached to generated code.}]
def appendOne (xs : List Nat) (x : Nat) : List Nat :=
  xs ++ [x]

def AppendOneSpec (xs : List Nat) (x : Nat) (out : List Nat) : Prop :=
  out.length = xs.length + 1

theorem appendOne_length (xs : List Nat) (x : Nat) :
  AppendOneSpec xs x (appendOne xs x) := by
  unfold AppendOneSpec appendOne
  simp
\end{lstlisting}

\subsection{Task Family Proof Patterns}

The following examples show why theorem generation is task dependent. The code blocks are schematic proof targets. Some include \texttt{sorry} because they illustrate the obligations BRIDGE discovers rather than completed verification certificates.

\paragraph{Sorting and searching.}
Sliding window and sorting tasks often require both achievability and optimality statements.

\begin{lstlisting}[style=colorfulstyle, language=Lean, caption={Maximum Beauty theorem targets.}]
def validSubseq (nums : List Int) (idxs : List (Fin nums.length)) : Prop :=
  idxs.Pairwise (· < ·)

def canTransform (nums : List Int) (k : Int)
    (idxs : List (Fin nums.length)) : Prop :=
  ∃ c : Int, ∀ i ∈ idxs, Int.natAbs (nums.get i - c) ≤ k.natAbs

theorem maximumBeauty_achievable (nums : List Int) (k : Int) :
  ∃ idxs : List (Fin nums.length),
    validSubseq nums idxs ∧
    canTransform nums k idxs ∧
    idxs.length = maximumBeauty nums k := by
  sorry

theorem maximumBeauty_optimal (nums : List Int) (k : Int) :
  ∀ idxs : List (Fin nums.length),
    validSubseq nums idxs →
    canTransform nums k idxs →
    idxs.length ≤ maximumBeauty nums k := by
  sorry
\end{lstlisting}

\paragraph{Combinatorial counting.}
Counting tasks require an exact correspondence between the implementation and a finite set of valid objects.

\begin{lstlisting}[style=colorfulstyle, language=Lean, caption={Combinatorial theorem targets.}]
def validRemovals (nums : Array Int) : Finset (Nat × Nat) :=
  let n := nums.size
  let pairs := (Finset.range (n + 1)).product (Finset.range (n + 1))
  pairs.filter (fun p => isValidRemoval nums p.1 p.2)

theorem incremovable_exact_count (nums : Array Int) :
  incremovableSubarrayCount nums = (validRemovals nums).card := by
  sorry

theorem incremovable_upper_bound (nums : Array Int) :
  incremovableSubarrayCount nums ≤ (nums.size + 1) ^ 2 := by
  sorry
\end{lstlisting}

\paragraph{Tree dynamic programs.}
Tree algorithms often expose the hardest gap between algorithm discovery and Lean accepted verification. Models may find the right recursive decomposition, but still fail to provide a termination measure acceptable to Lean.

\begin{lstlisting}[style=colorfulstyle, language=Lean, caption={Tree dynamic programming theorem target.}]
theorem dfs_correct
    (adj : Array (Array Nat)) (val : Array Int) (root : Nat) :
  let p := dfs adj val root root
  p.1 ≥ 0 ∧ p.2 ≥ 0 ∧ p.1 = optimalScore adj val root := by
  sorry
\end{lstlisting}

These examples illustrate why theorem/proof artifacts are useful diagnostics. They identify the mathematical obligations that a future prover or tool using agent must discharge, even when the current system cannot complete the full semantic proof.

\section{SFT and Training Details}
\label{app:sft_details}

This section gives the supervised fine tuning details behind the learnability results. The purpose is to test whether BRIDGE style reasoning can be internalized by a model, not to establish a scaling law.

\subsection{Training Data}

We constructed two matched 2k row OpenCodeInstruct style datasets from the same task distribution. The direct dataset pairs problem prompts with direct Lean generations. The functional dataset pairs analogous prompts with functional intermediate reasoning followed by the final Lean answer. Both datasets use the same formatting, filtering rules, train and validation split policy, and held out evaluation protocol.

\subsection{Training Setup}

Both variants fine tune \textsc{Goedel-Code-Prover-8B} with LoRA on one H100. We use LoRA rank 32, LoRA alpha 16, maximum sequence length 2048, learning rate $10^{-4}$, per device batch size 1, gradient accumulation 8, bf16 precision, seed 42, and 30 total epochs. The direct and functional runs use the same base checkpoint and are evaluated on held out BRIDGE suite, \textsc{VERINA}, and \textsc{CLEVER} tasks.

\begin{table}[!ht]
\centering
\caption{Goedel-Code-Prover-8B SFT results. Functional trace SFT improves over matched direct trace SFT on public benchmarks, with the strongest gain on \textsc{VERINA}.}
\label{tab:goedel_sft_summary}
\small
\setlength{\tabcolsep}{4pt}
\begin{tabular}{llccc}
\toprule
\textbf{Benchmark} & \textbf{Setting} & \textbf{Direct} & \textbf{Functional} & \textbf{$\Delta$} \\
\midrule
\textsc{VERINA} & Base Goedel pass@1 & 14/187 & 12/187 & -1.1 pp \\
\textsc{VERINA} & SFT e30 pass@1 & 38/187 & 55/187 & +9.1 pp \\
\textsc{CLEVER} & Base Goedel larger budget & 66/161 & 70/161 & +2.5 pp \\
\textsc{CLEVER} & SFT e30 larger budget & 30/161 & 33/161 & +1.9 pp \\
\bottomrule
\end{tabular}
\end{table}

\subsection{Operational Alignment}

Functional SFT also changes the shape of generated Lean code. On BRIDGE suite pass@5 generations, functional SFT outputs are shorter and cleaner than direct SFT outputs: 514.4 average extracted Lean characters versus 833.2, zero detected imports versus 876, zero \texttt{sorry}s versus 7, and 17 mutation or loop like artifacts versus 128. On \textsc{VERINA}, functional SFT is less imperative, with 5 detected mutation or loop like artifacts versus 40, and is substantially more accurate. On \textsc{CLEVER}, functional attempts solve more tasks, use fewer attempts, exhaust fewer budgets, and contain no detected imports, \texttt{sorry}s, or imperative artifacts.

\section{Agentic and Robustness Analyses}
\label{app:extended}

Here we collect supporting analyses that are useful for interpretation but too detailed for the main text.

\subsection{Tool Enabled Agent Sanity Check}
\label{app:codex_agent}

We ran a separate sanity check with a tool enabled Codex/GPT-5.2 workflow on the BRIDGE suite. This setting differs from the fixed budget prompting experiments: the agent can iteratively edit files, inspect Lean errors, run tests, and use tool feedback until the task is solved. Under this much stronger interaction model, all 178 tasks were solved in direct, imperative, and functional styles, so we do not present it as a fixed budget baseline.

The experiment is useful for positioning. It shows that the benchmark is not intrinsically impossible for a strong tool using agent. It also supports the paper's framing that BRIDGE is not only about final test pass rate; its value is in structuring the artifacts that an agent or prover manipulates. In these runs, functional style workflows reached completion about $4\times$ faster than direct workflows and produced better specification and theorem drafts under a combined LLM judge and human review rubric.

\subsection{Temperature Robustness}

\begin{figure}[!ht]
    \centering
    \includegraphics[width=0.85\linewidth]{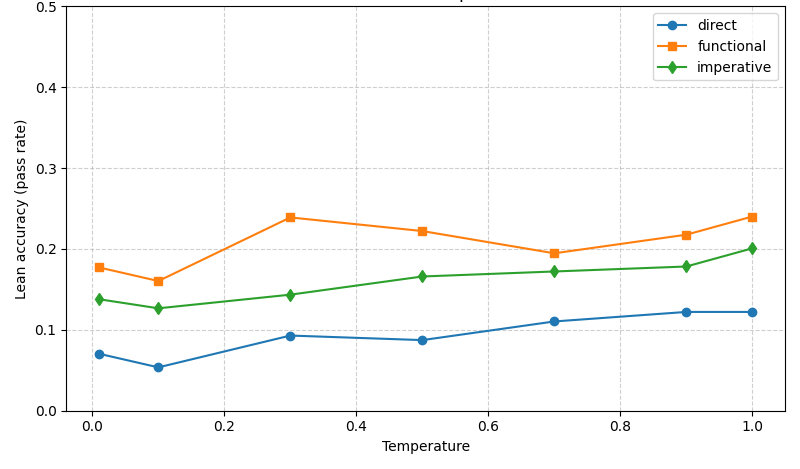}
\caption{Temperature robustness of reasoning paradigms. Pass@5 accuracy of \textsc{Claude Sonnet 4} as a function of sampling temperature for direct Lean, imperative Python bridge, and functional OCaml bridge prompting.}
    \label{fig:temperature_robustness}
\end{figure}

Functional prompting remains consistently stronger across temperatures, suggesting that the effect is not merely a sampling artifact. It remains useful across moderate changes in decoding randomness.

\subsection{BRIDGE Inference Time Framework}

\begin{figure}[!ht]
    \centering
    \includegraphics[width=\linewidth]{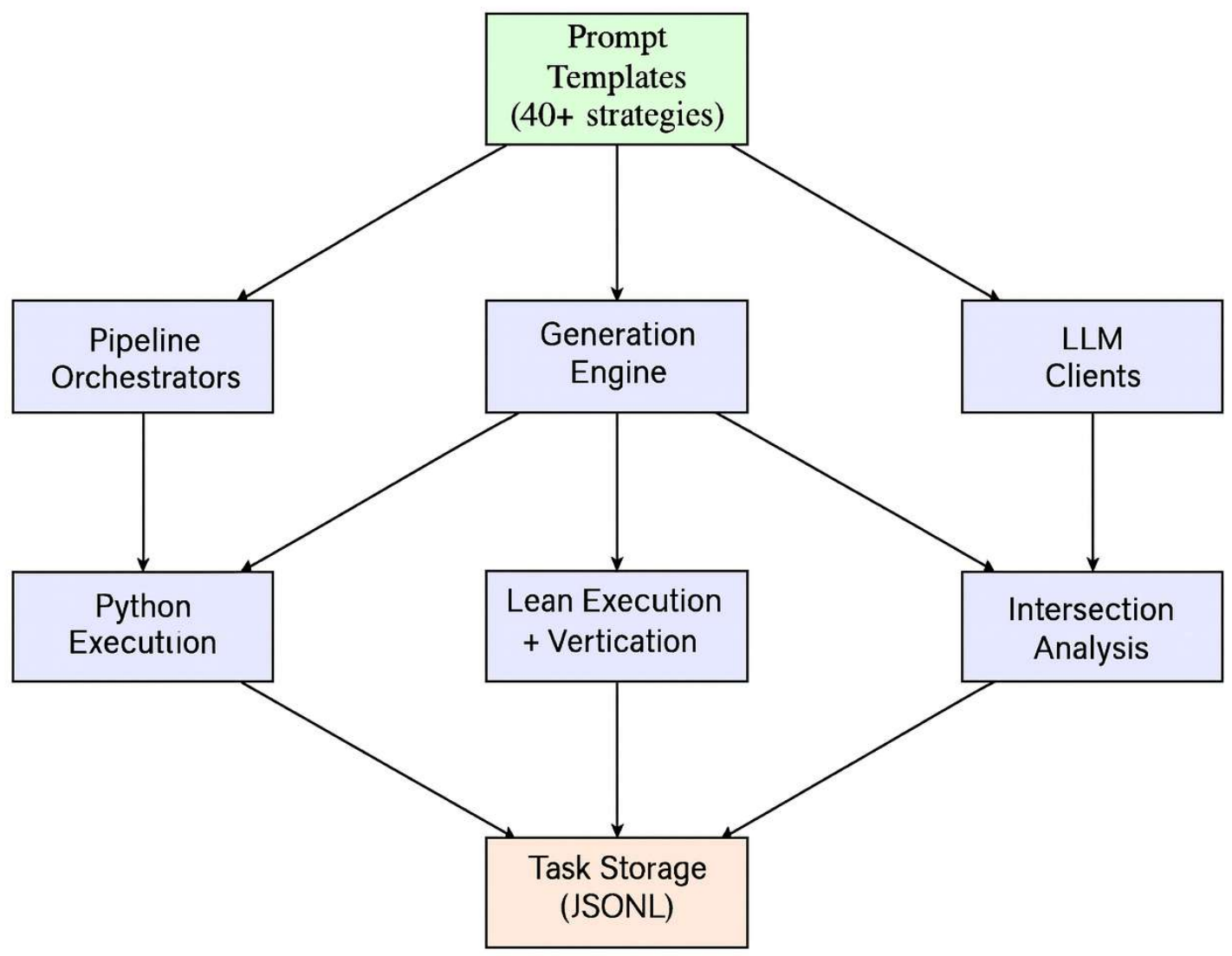}
    \caption{BRIDGE inference time framework. Code domain scaffolds produce Lean implementations, Specification domain prompts derive contracts and invariants, Theorem/Proof domain prompts generate correctness claims and proof attempts, and verifier feedback can be used for refinement.}
    \label{fig:system_architecture}
\end{figure}

Figure~\ref{fig:system_architecture} gives the full inference time view of BRIDGE. The main text presents the compact methodology view; this figure makes explicit where code, specifications, theorem statements, proof attempts, and verifier feedback enter the loop.

\subsection{Error Informed Python Code Refinement}

\begin{figure}[!ht]
    \centering
    \includegraphics[width=0.95\linewidth]{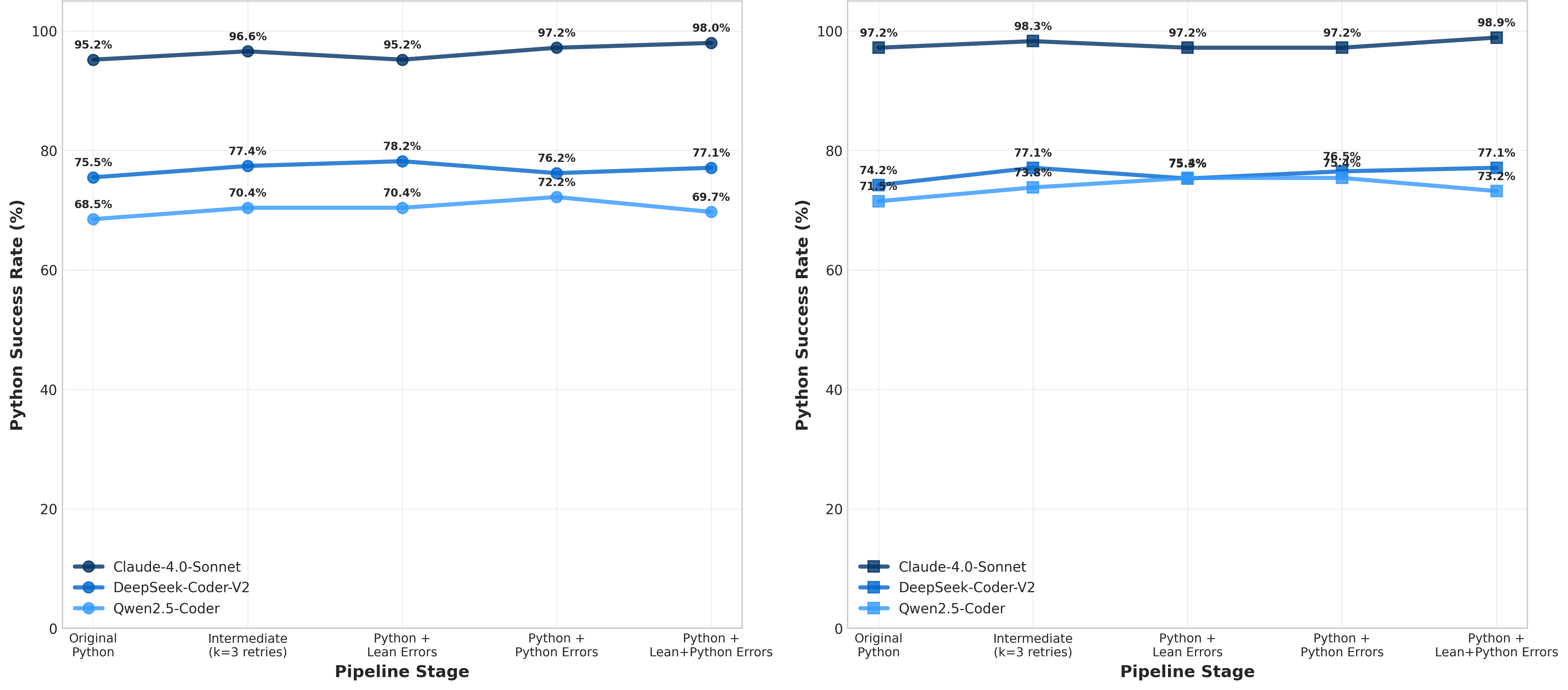}
    \caption{Python code success rates across refinement stages. Functional paradigms outperform imperative ones, and incorporating Lean error feedback improves upstream Python generation, although final Lean executable success remains the bottleneck.}
    \label{fig:python_lean_improvement}
\end{figure}

We also study a pipeline where Python generation benefits from retry mechanisms and cross domain error signals from Lean. Lean derived errors can improve Python code quality even when Lean translation ultimately fails. The gap between high Python success and lower Lean executable success highlights the central difficulty: the bottleneck is often translating an algorithm into a total, typed, Lean accepted definition.

\subsection{Failure Patterns}

Functional scaffolds reduce several common Lean failure modes. Direct prompting often fails through syntax errors, missing type annotations, partial functions, and loops that do not translate cleanly into Lean. Imperative scaffolds can recover algorithmic intent, but they often require the model to invent loop invariants and state transition arguments after the fact. Functional scaffolds instead bias the model toward recursive definitions, immutable data, and compositional structure, which are closer to Lean's accepted programming style.

The most important failure classes are syntax failures, type failures, and termination or totality failures. Functional reasoning reduces syntax and type failures relative to direct generation, and it tends to produce more structurally recursive candidates. This does not eliminate termination errors, especially for graph and tree algorithms, but it makes the generated definitions closer to the form Lean can check.

\subsection{Theorem Reasoning as Code Guidance}

We also evaluate whether theorem oriented reasoning can improve code generation before proof search. A theorem first strategy reaches higher pass@5 than direct Lean generation in a subset of tasks, especially for mathematically heavy problems such as geometry, combinatorial counting, and bit manipulation. When both theorem guided and direct strategies succeed, their final algorithms are often similar, suggesting that theorem reasoning helps the model find the same core algorithm through a more constrained route. This supports the BRIDGE view that theorem/proof artifacts serve two roles: they identify formal obligations, and they can guide implementation search.

\section{Additional Related Work}
\label{app:related_work}

\paragraph{Large scale formal verification.}
Large scale verification projects show that full functional correctness proofs are possible for realistic systems, but they require extensive expert effort. Systems such as the \textsc{seL4} microkernel and the \textsc{CompCert} verified compiler demonstrate the strength of machine checked proofs, while also illustrating why scalable formal verification remains difficult in practice. Solver based tools such as \textsc{Dafny}, \textsc{Boogie}, \textsc{Why3}, and \textsc{Verus}~\cite{zhu2023verus} lower this burden by translating annotated programs into verification conditions that can often be discharged automatically. These systems motivate BRIDGE's focus on the artifacts that appear before automation succeeds: executable code, formal specifications, theorem statements, and proof attempts must be structured well enough for a verifier or prover to use them.

\paragraph{Solver based verification with LLMs.}
Solver based verification is often more forgiving than interactive theorem proving because many local proof obligations can be handled automatically once the right annotations and specifications are present. Benchmarks such as \textsc{DafnyBench} and systems such as \textsc{AutoSpec} show that LLMs can perform well in this setting, especially when the task is to produce code and contracts that an SMT backend can finish~\cite{loughridge2024dafnybench,wen2024autospec}. \textsc{Clover} uses a closed loop pipeline in which an LLM generates code, documentation, and annotations, then applies consistency checks and solver based verification to decide acceptance~\cite{sun2024clover}. These results suggest that when the backend can close the proof gap, the main difficulty shifts toward generating correct, complete, and nonvacuous specifications. BRIDGE complements this line of work by studying how the reasoning scaffold changes the generated code and formal artifacts before the solver or prover takes over.

\paragraph{Specification synthesis and intent alignment.}
Recent work has begun to treat formal specification generation as a first class task rather than a byproduct of code generation. \textsc{VeriSpecGen} targets Lean specification synthesis on \textsc{VERINA} by decomposing natural language intent into atomic requirements, generating requirement targeted tests, and using traceability maps to localize and repair failed clauses~\cite{ye2026verispecgen}. This is closely related to BRIDGE's Specification domain. The difference is emphasis: specification synthesis systems focus on improving the specification itself, while BRIDGE studies how specifications interact with executable code, theorem statements, and proof attempts through a shared reasoning scaffold. In particular, BRIDGE evaluates specifications not only by elaboration, but also by nonvacuity, correct implementation acceptance, wrong implementation rejection, and downstream usefulness.

\paragraph{Interactive theorem proving with LLMs.}
Interactive theorem provers expose sharper limits for LLM based verification. Public Lean benchmarks such as \textsc{VERINA} and \textsc{CLEVER} show that models can generate plausible code, specifications, theorem statements, and proof attempts, but still struggle to connect them into reliable verified artifacts~\cite{ye2025verina,clever2025}. In these settings, success requires more than local annotation generation: the implementation must be expressible in Lean, the specification must capture the intended behavior, the theorem statement must connect the two, and the proof must be accepted by the kernel. Prior benchmark results show that code and specification generation are substantially easier than full proof completion, and that performance drops sharply when proofs must compose across richer artifacts. BRIDGE is designed around this bottleneck. Rather than treating proof failure as an isolated prover problem, it treats the entire code, specification, theorem, and proof chain as an artifact alignment problem.

\paragraph{Compositional and cross system verification benchmarks.}
Recent benchmarks also show that local verification success does not automatically scale to larger verification tasks. \textsc{DAFNYCOMP} studies compositional specification problems in Dafny, while the \textsc{Vericoding Benchmark} studies verification oriented coding across systems~\cite{dafnycomp2025,vericoding2025}. These settings emphasize that even when an LLM can generate locally plausible code or annotations, composing specifications and proofs across modules remains difficult. This limitation is directly relevant to BRIDGE. The present work focuses on the front end of verified synthesis and mostly algorithmic tasks, but the same artifact alignment problem becomes more important in long horizon, multi module, or cross system verification settings.

\paragraph{Multi view program representations.}
A related line of work represents programs through multiple semantic views, including source code, natural language descriptions, syntax trees, data flow, documentation, and generated explanations. Models such as \textsc{CodeBERT} and \textsc{GraphCodeBERT} show that combining code with structural or textual views can improve program understanding and generation~\cite{feng2020codebert,guo2021graphcodebert}. BRIDGE specializes this multi view idea to formal verification. The views are not only code and text, but executable implementations, specifications, theorem statements, and proof attempts. Unlike ordinary program representations, these artifacts must survive formal checks and remain mutually meaningful inside a proof assistant or solver backed verification system.

\section{Comprehensive Prompt System for BRIDGE}
\label{app:prompt_system}
\label{app:F}

Our \textsc{BRIDGE} framework uses a modular prompt system to evaluate structured reasoning approaches across the three domains of formal verification. Over 40 strategies are organized into three families: \emph{Code Domain} (Lean implementations), \emph{Specification Domain} (formal contracts), and \emph{Theorem/Proof Domain} (mathematical property discovery and proof attempts). We present unified templates with strategy-specific content marked as \texttt{[SWAP]} placeholders.

\paragraph{Prompt Construction Procedure.}
To construct a complete prompt from a template, we follow a systematic substitution process. First, the natural language problem description (from the LeetCode dataset) is prepended to the template. Next, \texttt{[SWAP:STRATEGY]} is replaced with one of the strategy options listed in the template (e.g., ``Direct'', ``Python Bridge'', ``Haskell Functional''). The \texttt{[SWAP:REASONING\_STYLE]} placeholder is replaced with the corresponding reasoning approach name (e.g., ``Functional Reasoning'' for Haskell, ``Imperative Reasoning'' for C++). The \texttt{[SWAP:REASONING\_DETAILS]} placeholder is filled with strategy-specific reasoning guidelines provided in the subsequent template boxes. Metadata placeholders such as \texttt{InputType}, \texttt{OutputType}, and \texttt{\{\{ function\_name \}\}} are extracted from the problem statement: \texttt{InputType} and \texttt{OutputType} are inferred from the problem's input/output specifications, while \texttt{\{\{ function\_name \}\}} is derived from the problem's function signature or a canonical name based on the problem description. All placeholders are replaced before sending the prompt to the model, ensuring complete and consistent prompt construction across all strategies.

\subsection{Code Domain Strategies (Lean)}

The Code Domain tests our hypothesis that reasoning strategies affect \textsc{Lean} generation success. We use a unified template with 6 reasoning strategies:

\begin{promptboxICML}{BRIDGE Code Domain Template}
\small
\textbf{Instructions.}
Solve the task by applying the reasoning paradigm in
\texttt{[SWAP:STRATEGY]} and translate the solution to \textsc{Lean}.

\medskip
\textbf{\texttt{[SWAP:STRATEGY]} options:}
\begin{itemize}[leftmargin=1em, itemsep=0.25em]
  \item \textbf{Direct}: Solve directly in \textsc{Lean}
  \item \textbf{Python Bridge}: Reason in Python logic first, then translate
  \item \textbf{Haskell Functional}: Pattern matching, recursion, type-driven development
  \item \textbf{OCaml Type-Guided}: Type safety and mathematical invariants
  \item \textbf{Double Lean}: Mathematical reasoning in Lean style, then implement
  \item \textbf{C++ Imperative}: Procedural logic, then translate to functional \textsc{Lean}
\end{itemize}

\medskip
\texttt{Step 1: [SWAP:REASONING\_STYLE]} —
\texttt{[SWAP:REASONING\_DETAILS]}

\textbf{Note:} \texttt{[SWAP:REASONING\_STYLE]} is replaced with the strategy name (e.g., ``Functional Reasoning'', ``Imperative Reasoning'') corresponding to the selected \texttt{[SWAP:STRATEGY]} option.

\smallskip
\texttt{Step 2: Lean Translation}

\begin{lstlisting}[style=verbatimish, language=Lean]
import Std
import Mathlib

def solution (input : InputType) : OutputType :=
  -- Implementation from [SWAP:STRATEGY]
  sorry
\end{lstlisting}
\end{promptboxICML}

\begin{promptboxICML}{BRIDGE-I (C++ / Python)}
\small
\textbf{\texttt{[SWAP:REASONING\_DETAILS]} for C++:}
\begin{itemize}[leftmargin=1em, itemsep=0.25em]
  \item Loops, arrays, pointers, algorithmic thinking
  \item Data structures (vectors, maps, sets)
  \item Memory management, efficiency, complexity
\end{itemize}

\medskip
\textbf{\texttt{[SWAP:REASONING\_DETAILS]} for Python:}
\begin{itemize}[leftmargin=1em, itemsep=0.25em]
  \item Python-like logic and built-ins
  \item Simple, readable solutions
\end{itemize}
\end{promptboxICML}

\begin{promptboxICML}{BRIDGE (Haskell / OCaml / Double Lean)}
\small
\textbf{\texttt{[SWAP:REASONING\_DETAILS]} for Haskell:}
\begin{itemize}[leftmargin=1em, itemsep=0.25em]
  \item Pattern matching and structural recursion
  \item Functional composition and list processing
  \item Type signatures and algebraic data types
\end{itemize}

\medskip
\textbf{\texttt{[SWAP:REASONING\_DETAILS]} for OCaml:}
\begin{itemize}[leftmargin=1em, itemsep=0.25em]
  \item Immutable data structures and recursion
  \item Algebraic data types, variants, records
  \item Tail recursion and compositional design
\end{itemize}

\medskip
\textbf{\texttt{[SWAP:REASONING\_DETAILS]} for Double Lean:}
\begin{itemize}[leftmargin=1em, itemsep=0.25em]
  \item Mathematical properties and invariants
  \item Inductive definitions and recursive structures
  \item Proof strategies grounded in type theory
\end{itemize}
\end{promptboxICML}

\subsection{Specification Domain Strategies (Python)}

\begin{promptboxICML}{Unified Specification Domain Template}
\small
\textbf{Instructions.}
Solve the problem using \texttt{[SWAP:STRATEGY]} reasoning, then implement the
solution in Python with formal contracts.

\medskip
\textbf{\texttt{[SWAP:STRATEGY]} options:}
\begin{itemize}[leftmargin=1em, itemsep=0.25em]
  \item \textbf{Direct}: Write Python solution directly without intermediate reasoning
  \item \textbf{Design by Contract}: Define preconditions, postconditions, and invariants
  \item \textbf{Dafny-style}: Use requires/ensures clauses and verification conditions
  \item \textbf{Property-Based Testing}: Identify universal mathematical properties
  \item \textbf{Functional Programming}: Use immutable data and pure functions
  \item \textbf{Defensive Programming}: Validate inputs and handle errors robustly
  \item \textbf{Algorithmic Thinking}: Decompose into algorithmic subproblems
  \item \textbf{Test-Driven Development}: Design by reasoning about test cases first
\end{itemize}

\medskip
\texttt{Step 1: [SWAP:REASONING\_APPROACH]} —
Apply strategy-specific reasoning.

\smallskip
\texttt{Step 2: Python Implementation} —
Translate reasoning into Python with formal contracts.

\begin{lstlisting}[style=verbatimish, language=Python]
# APPROACH:
# First reason using [SWAP:STRATEGY] principles,
# then implement in Python.

# Solution Process
# Step 1: [SWAP:STRATEGY] Reasoning
[SWAP:DETAILED_REASONING_GUIDELINES]

# Step 2: Python Implementation
# Apply [SWAP:STRATEGY] reasoning to create robust code with contracts

[SWAP:CONTRACT_DECORATORS]
def function_name(parameters):
    """
    [SWAP:DOCSTRING_WITH_CONTRACTS]
    """
    # Implementation guided by [SWAP:STRATEGY] reasoning
    pass
\end{lstlisting}
\end{promptboxICML}

\subsubsection{Concrete Prompt Examples for Each Strategy}
\begin{promptboxICML}{Prompt — Design by Contract Reasoning}
\small
\textbf{Reasoning Focus.}
Define rigorous preconditions, postconditions, and invariants that govern
function behavior. This approach emphasizes formal contracts that specify
exactly what conditions must hold before and after execution.

\begin{lstlisting}[style=verbatimish, language=Python]
import deal
from typing import List

@deal.pre(lambda data: isinstance(data, list) and len(data) > 0)
@deal.pre(lambda data: all(isinstance(x, int) for x in data))
@deal.post(lambda result: result >= 0)
@deal.ensure(lambda data, result: result <= sum(x for x in data if x > 0))
def contract_solution(data: List[int]) -> int:
    """
    Precondition: non-empty list of integers
    Postcondition: result >= 0 and bounded by positive sum
    Invariant: input data structure remains unchanged
    """
    assert isinstance(data, list) and len(data) > 0
    assert all(isinstance(x, int) for x in data)
    result = sum(x for x in data if x >= 0)
    assert result >= 0
    return result
\end{lstlisting}
\end{promptboxICML}

\begin{promptboxICML}{Prompt — Dafny-Style Specification Reasoning}
\small
\textbf{Reasoning Focus.}
Apply formal verification principles with requires/ensures clauses and
mathematical assertions, inspired by the Dafny verification language.

\begin{lstlisting}[style=verbatimish, language=Python]
import deal
from typing import List

@deal.chain(
    deal.pre(lambda data: all(isinstance(x, int) for x in data)),
    deal.pre(lambda data: data == sorted(data)),
    deal.post(lambda result: result == sorted(result)),
    deal.ensure(lambda data, result: all(x >= 0 for x in result))
)
def dafny_style_solution(data: List[int]) -> List[int]:
    """
    Requires: input sorted list of ints
    Ensures: output sorted and non-negative
    Invariant: sorted order maintained throughout
    """
    assert all(isinstance(x, int) for x in data)
    assert data == sorted(data)
    result = [abs(x) for x in data]
    assert result == sorted(result)
    return result
\end{lstlisting}
\end{promptboxICML}

\begin{promptboxICML}{Prompt — Property-Based Testing Reasoning}
\small
\textbf{Reasoning Focus.}
Identify universal mathematical properties that must hold across all valid
input domains, emphasizing algebraic invariants and bounds.

\begin{lstlisting}[style=verbatimish, language=Python]
import deal
from typing import List

@deal.chain(
    deal.pre(lambda data: isinstance(data, list)),
    deal.post(lambda result: result >= 0),
    deal.ensure(lambda data, result: result <= sum(abs(x) for x in data))
)
def property_solution(data: List[int]) -> int:
    """
    Universal Properties:
    - Commutativity: order independence
    - Monotonicity: adding positives increases result
    - Non-negativity: result >= 0
    """
    return sum(x for x in data if x >= 0)
\end{lstlisting}
\end{promptboxICML}

\begin{promptboxICML}{Prompt — Test-Driven Development Reasoning}
\small
\textbf{Reasoning Focus.}
Design solutions by reasoning about comprehensive test cases first, following
the Red--Green--Refactor cycle.

\begin{lstlisting}[style=verbatimish, language=Python]
import deal
from typing import List

@deal.chain(
    deal.pre(lambda data: isinstance(data, list)),
    deal.post(lambda result: isinstance(result, int) and result >= 0)
)
def tdd_solution(data: List[int]) -> int:
    """
    TDD Design Process:
    - Red: define failing tests
    - Green: minimal implementation
    - Refactor: optimize safely
    """
    if not isinstance(data, list) or len(data) == 0:
        return 0

    positive_sum = sum(x for x in data if isinstance(x, int) and x > 0)
    return positive_sum

def comprehensive_test_suite():
    assert tdd_solution([1, 2, 3]) == 6
    assert tdd_solution([]) == 0
    assert tdd_solution([-1, -2]) == 0
    assert tdd_solution([1, "2", None]) == 1
\end{lstlisting}
\end{promptboxICML}

\begin{promptboxICML}{Prompt — Defensive Programming Reasoning}
\small
\textbf{Reasoning Focus.}
Implement comprehensive input validation, error handling, and fail-safe
behaviors to ensure robustness.

\begin{lstlisting}[style=verbatimish, language=Python]
import deal
import logging
from typing import List, Union

@deal.chain(
    deal.safe,
    deal.post(lambda result: isinstance(result, int) and result >= 0)
)
def defensive_solution(data: Union[List[int], None]) -> int:
    """
    Defensive Strategies:
    - Validate inputs
    - Handle malformed data gracefully
    - Provide safe fallback behavior
    """
    if data is None or not isinstance(data, list):
        logging.warning("Invalid input, using safe default")
        return 0

    try:
        valid_ints = [x for x in data if isinstance(x, int)]
        return sum(x for x in valid_ints if x >= 0)
    except Exception as e:
        logging.error(f"Computation error: {e}")
        return 0
\end{lstlisting}
\end{promptboxICML}
\begin{promptboxICML}{Prompt — Algorithmic Thinking Reasoning}
\small
\textbf{Reasoning Focus.}
Systematically decompose problems into algorithmic subcomponents, emphasizing
structure, correctness, and complexity analysis.

\begin{lstlisting}[style=verbatimish, language=Python]
import deal
from typing import List

@deal.chain(
    deal.pre(lambda data: isinstance(data, list)),
    deal.post(lambda result: isinstance(result, int) and result >= 0)
)
def algorithmic_solution(data: List[int]) -> int:
    """
    Algorithm Design:
    1. Preprocessing
    2. Core computation
    3. Post-processing validation

    Complexity: O(n) time, O(1) space
    """
    if not data:
        return 0

    result = sum(x for x in data if isinstance(x, int) and x > 0)
    return max(0, result)
\end{lstlisting}
\end{promptboxICML}

\subsection{Theorem/Proof Domain Strategies}

The Theorem/Proof Domain uses five pathways for robust mathematical property discovery, theorem generation, and bounded proof attempts.
Each pathway approaches mathematical property identification from different perspectives to ensure comprehensive coverage.
\begin{promptboxICML}{Unified Theorem Statements Template}
\small
\textbf{\texttt{[SWAP:PATHWAY]} options:}
\begin{itemize}[leftmargin=1em, itemsep=0.25em]
  \item \textbf{Natural Language}: Extract properties directly from the problem description
  \item \textbf{Unit Tests}: Generalize mathematical claims from test patterns
  \item \textbf{Code Analysis}: Reverse-engineer properties from implementation structure
  \item \textbf{Type-Guided}: Leverage refined types for precise theorem generation
  \item \textbf{Termination}: Focus on termination proofs and complexity bounds
\end{itemize}

\begin{lstlisting}[style=verbatimish, language=Lean]
-- Task: Generate Lean implementation AND theorems via [SWAP:PATHWAY]

-- PATHWAY:
-- [SWAP:PATHWAY_SPECIFIC_ANALYSIS]

-- Your task is to create:
-- 1) Lean Implementation: complete solution to the problem
-- 2) Formal Theorems: mathematical properties discovered via [SWAP:PATHWAY]

import Std
import Mathlib

def solution (input : InputType) : OutputType :=
  -- Implementation guided by [SWAP:PATHWAY] analysis
  sorry

-- [SWAP:THEOREM_TYPE]
theorem [SWAP:THEOREM_NAME] (input : InputType) :
  [SWAP:THEOREM_STATEMENT] := by
  sorry
\end{lstlisting}
\end{promptboxICML}

\subsubsection{Concrete Pathway Examples}
\begin{promptboxICML}{Prompt — Natural Language $\to$ Lean Theorems}
\small
\textbf{Analysis Focus.}
Extract key mathematical properties and correctness conditions from the natural
language statement and express them as Lean theorems about
\texttt{\{\{ function\_name \}\}}. Implement \texttt{\{\{ function\_name \}\}}
and generate \emph{Theorem 1} (functional correctness) and \emph{Theorem 2}
(input/output relationship).

\begin{lstlisting}[style=verbatimish, language=Lean]
-- Theorem 1: functional correctness
theorem {{ function_name }}_correctness
    ({{ function_params }}) (h : precondition {{ function_params }}) :
    property ({{ function_name }} {{ param_names }}) := by
  sorry

-- Theorem 2: input/output relationship
theorem {{ function_name }}_input_output_relationship
    ({{ function_params }}) :
    relation {{ param_names }} ({{ function_name }} {{ param_names }}) := by
  sorry
\end{lstlisting}
\end{promptboxICML}
\begin{promptboxICML}{Prompt — Unit Test Pattern $\to$ Lean Theorems}
\small
\textbf{Analysis Focus.}
Study unit test patterns to infer relationships and invariants they enforce,
then express these as Lean theorems about \texttt{\{\{ function\_name \}\}}.
Implement \texttt{\{\{ function\_name \}\}} and generate \emph{Theorem 1}
(a property generalized from tests) and \emph{Theorem 2} (an invariant relating
multiple calls).

\begin{lstlisting}[style=verbatimish, language=Lean]
-- Theorem 1: property generalized from test patterns
theorem {{ function_name }}_test_pattern_theorem
    ({{ function_params }}) :
    property_from_tests {{ param_names }}
      ({{ function_name }} {{ param_names }}) := by
  sorry

-- Theorem 2: invariant identified from tests
theorem {{ function_name }}_test_invariant_theorem
    (input1 input2 : {{ input_type }}) :
    invariant_property
      ({{ function_name }} input1)
      ({{ function_name }} input2) := by
  sorry
\end{lstlisting}
\end{promptboxICML}

\begin{promptboxICML}{Prompt — Code Implementation $\to$ Lean Theorems}
\small
\textbf{Analysis Focus.}
Reverse-engineer mathematical properties from an existing Lean implementation
of \texttt{\{\{ function\_name \}\}}. Derive theorems that verify algorithmic
correctness and implementation-specific invariants.

\begin{lstlisting}[style=verbatimish, language=Lean]
-- Theorem 1: functional correctness of {{ function_name }}
theorem {{ function_name }}_correctness
    ({{ function_params }}) (h : valid_input {{ param_names }}) :
    satisfies_specification
      ({{ function_name }} {{ param_names }}) {{ param_names }} := by
  sorry

-- Theorem 2: mathematical property preservation
theorem {{ function_name }}_property_preserved
    ({{ function_params }}) :
    property_holds {{ param_names }}
      ({{ function_name }} {{ param_names }}) := by
  sorry

-- Theorem 3: algorithmic invariant
theorem {{ function_name }}_algorithm_invariant
    ({{ function_params }}) :
    forall_steps_satisfy_invariant
      ({{ function_name }} {{ param_names }}) := by
  sorry
\end{lstlisting}
\end{promptboxICML}

\begin{promptboxICML}{Prompt — Termination $\to$ Complexity Proofs}
\small
\textbf{Analysis Focus.}
Generate Lean theorems about termination and computational complexity for
\texttt{\{\{ function\_name \}\}}. Implement \texttt{\{\{ function\_name \}\}}
with termination in mind and produce termination-focused statements.

\begin{lstlisting}[style=verbatimish, language=Lean]
-- Theorem 1: termination of {{ function_name }}
theorem {{ function_name }}_terminates
    ({{ function_params }}) :
    ∃ output : {{ return_type }},
      {{ function_name }} {{ param_names }} = output := by
  sorry

-- Theorem 2: well-founded recursion (if applicable)
theorem {{ function_name }}_recursion_well_founded
    ({{ function_params }}) :
    WellFounded (recursion_relation {{ param_names }}) := by
  sorry

-- Theorem 3: complexity bound for {{ function_name }}
theorem {{ function_name }}_complexity_bound
    ({{ function_params }}) :
    computation_steps ({{ function_name }} {{ param_names }}) ≤
      complexity_function (size {{ param_names }}) := by
  sorry
\end{lstlisting}
\end{promptboxICML}
\begin{promptboxICML}{Prompt — Type-Guided $\to$ Lean Theorems}
\small
\textbf{Analysis Focus.}
Given a strongly typed Lean implementation of \texttt{\{\{ function\_name \}\}},
generate theorems that leverage refined types and mathematical structure to
prove stronger correctness and invariants.

\begin{lstlisting}[style=verbatimish, language=Lean]
-- Theorem 1: type-level correctness for {{ function_name }}
theorem {{ function_name }}_type_level_correctness
    ({{ function_params }}) :
    TypeConstraintSatisfied
      ({{ function_name }} {{ param_names }}) := by
  sorry

-- Theorem 2: preservation of mathematical structure
theorem {{ function_name }}_structure_preservation
    ({{ function_params }}) :
    PreservesStructure
      ({{ function_name }} {{ param_names }})
      {{ param_names }} := by
  sorry

-- Theorem 3: enhanced invariant using refined types
theorem {{ function_name }}_enhanced_invariant
    ({{ function_params }}) :
    StrongerInvariant {{ param_names }}
      ({{ function_name }} {{ param_names }}) := by
  sorry
\end{lstlisting}
\end{promptboxICML}

\subsection{Iterative Improvement Strategies}

The framework implements iterative improvement mechanisms using chain-of-thought error analysis.
This enables systematic refinement of failed solutions through structured error diagnosis and targeted corrections.
\begin{promptboxICML}{Prompt — Chain-of-Thought Error Analysis}
\small
\textbf{Goal.}
Systematically improve failed solutions through structured error analysis and
reasoning refinement. The approach follows a four-phase methodology to identify,
analyze, and correct solution failures.

\medskip
\textbf{Process: Four Steps}
\begin{itemize}[leftmargin=1em, itemsep=0.25em]
  \item \textbf{Step 1: Error Classification.}
  Identify syntax errors, semantic errors, logic errors, type mismatches, and
  edge case failures.
  \item \textbf{Step 2: Root Cause Analysis.}
  Determine why the errors occurred and which concepts were misunderstood or
  mis-modeled.
  \item \textbf{Step 3: Solution Strategy Design.}
  Plan algorithmic changes, type corrections, edge case handling, and
  specification fixes.
  \item \textbf{Step 4: Systematic Implementation.}
  Apply improvements based on the analysis and re-verify correctness.
\end{itemize}

\begin{lstlisting}[style=verbatimish, language=Python]
# RETRY ATTEMPT {{ retry_attempt }} / {{ max_retries }}

# Your previous solution failed.
# Analyze the errors and generate an improved solution.

# Previous Python / Lean Code (FAILED)
{{ previous_python_or_lean_code }}

# Error Messages and Feedback
{{ python_or_lean_errors }}
\end{lstlisting}
\end{promptboxICML}

\begin{promptboxICML}{Meta-Analysis Process}
\small
\textbf{Analysis Methodology.}
Examine theorems generated across all five proof pathways to identify robust
mathematical properties. The process synthesizes pathway-specific insights into
integrated theorem sets with enhanced verification coverage.

\begin{lstlisting}[style=verbatimish, language=Python]
{
  "intersection_analysis": {
    "common_concepts": [
      "monotonicity",
      "bounds",
      "optimality",
      "correctness"
    ],
    "shared_properties": [
      "invariant_preservation",
      "edge_case_handling"
    ],
    "robust_theorems": [
      "most_frequently_appearing",
      "highest_mathematical_rigor"
    ],
    "pathway_specific_insights": [
      "unique_contributions_per_pathway"
    ]
  },
  "final_theorem_selection": [
    "solution_correctness: essential functional correctness property",
    "solution_bounds: mathematical bounds and constraint verification",
    "solution_termination: computational termination guarantees"
  ],
  "complete_Lean_file": "
import Std
import Mathlib

def solution (input : InputType) : OutputType :=
  complete_implementation

theorem solution\_bounds (input : InputType) :
  0 \\le solution input \\wedge solution input \\le upper\\_bound input := by
  sorry
  -- omitted for brevity
"
}
\end{lstlisting}
\end{promptboxICML}

\section{Theorem and Proof Examples and Analysis}
\label{app:proof_examples}

We present selected examples illustrating why generating good theorem/proof artifacts is hard and how BRIDGE addresses them. Each highlights how different algorithmic categories require distinct reasoning strategies, motivating our three-domain approach to formal verification. All code examples and theorem statements shown in this section were generated by \textsc{Claude Sonnet 4} using the BRIDGE framework; proof obligations are shown with \texttt{sorry} when they are intended targets rather than completed proofs.

\subsection{Sorting and Searching: Maximum Beauty}

The \texttt{maximumBeauty} problem exemplifies the sorting/searching category, where correctness depends on range constraints and sliding window invariants. The task is to find the longest subsequence whose elements can all be transformed to the same value within a tolerance $k$.

\begin{lstlisting}[style=colorfulstyle, language=Lean, caption={Maximum Beauty Implementation (truncated)}, label={lst:max_beauty_impl}]
def maximumBeauty (nums : List Int) (k : Int) : Int :=
  let arr := nums.toArray.qsort (· ≤ ·)
  -- crude but safe upper bound: each step increments l or r,
  -- and both are ≤ arr.size, so ≤ 2 * arr.size steps.
  let fuel0 := 2 * arr.size.succ

  -- structurally recursive on `fuel`
  let rec go : Nat → Nat → Nat → Nat → Int
  | 0,      l, r, best => best
  | fuel+1, l, r, best =>
      if h : r = arr.size then
        -- window has reached the end
        best
      else
        if arr[r]! - arr[l]! ≤ 2 * k then
          -- expand window to the right
          go fuel l (r + 1) (max best (r - l + 1))
        else
          -- shrink window from the left
          go fuel (l + 1) r best

  go fuel0 0 0 0

\end{lstlisting}

Verification relies on proving (i) that the algorithm returns an achievable subsequence, (ii) that it is optimal, and (iii) that the sliding window invariant is preserved:

\begin{lstlisting}[style=colorfulstyle, language=Lean, caption={Correctness Theorems}, label={lst:max_beauty_theorems}]
/-
  Specifications for subsequences and the "can be transformed" property.
  We model a subsequence by a list of indices `Fin nums.length`,
  in strictly increasing order.
-/

/-- `idxs` is a (strictly increasing) subsequence of the positions of `nums`. -/
def validSubseq (nums : List Int) (idxs : List (Fin nums.length)) : Prop :=
  idxs.Pairwise (· < ·)

/--
  There exists some center value `c` such that every chosen element
  is within `k` (in absolute value) of `c`.
-/
def canTransform (nums : List Int) (k : Int) (idxs : List (Fin nums.length)) : Prop :=
  ∃ c : Int, ∀ i ∈ idxs, |nums.get i - c| ≤ k

/--
  Correctness: `maximumBeauty` returns the size of *some* valid subsequence
  that can be transformed within radius `k`.
-/
theorem maximumBeauty_correct (nums : List Int) (k : Int) :
  ∃ idxs : List (Fin nums.length),
    validSubseq nums idxs ∧
    canTransform nums k idxs ∧
    idxs.length = maximumBeauty nums k := by
  sorry

/--
  Optimality: no valid transformable subsequence is longer than `maximumBeauty`.
-/
theorem maximumBeauty_optimal (nums : List Int) (k : Int) :
  ∀ idxs : List (Fin nums.length),
    validSubseq nums idxs →
    canTransform nums k idxs →
    idxs.length ≤ maximumBeauty nums k := by
  sorry
\end{lstlisting}

This domain benefits from functional reasoning because the recursive sliding window structure aligns naturally with structural recursion and proof mechanisms.

\subsection{Combinatorics: Incremovable Subarray Count}

Combinatorial problems stress counting correctness and exhaustive enumeration. The \texttt{incremovableSubarrayCount} problem counts subarrays whose removal leaves a strictly increasing sequence.

\begin{lstlisting}[style=colorfulstyle, language=Lean, caption={Incremovable Subarray Count (core loop)}, label={lst:incremovable_core}]

/-- Check if an array of `Int`s is strictly increasing. -/
def isStrictlyIncreasing (nums : Array Int) : Bool :=
  Id.run do
    let n := nums.size
    if n ≤ 1 then
      return true
    let mut ok := true
    let mut prev := nums[0]!
    for i in [1:n] do
      let x := nums[i]!
      if prev < x then
        prev := x
      else
        ok := false
    return ok

/--
`isValidRemoval nums i j` is true iff removing the subarray `nums[i .. j)`
(half-open interval) leaves a strictly increasing array.
We assume `0 ≤ i ≤ j ≤ nums.size`; otherwise we return `false`.
-/
def isValidRemoval (nums : Array Int) (i j : Nat) : Bool :=
  let n := nums.size
  if h : i ≤ j ∧ j ≤ n then
    let left  := nums.extract 0 i   -- [0, i)
    let right := nums.extract j n   -- [j, n)
    let remaining := left ++ right
    isStrictlyIncreasing remaining
  else
    false

/--
Count the number of subarrays whose removal leaves a strictly increasing sequence.
-/
def incremovableSubarrayCount (nums : Array Int) : Nat :=
  Id.run do
    let n := nums.size
    let mut count := 0
    for i in [0:n] do
      for j in [i:n] do
        if isValidRemoval nums i j then
          count := count + 1
    return count

\end{lstlisting}

Typical theorems ensure (i) exact counting of valid removals, (ii) a trivial size-squared upper bound, and (iii) a closed form for strictly increasing inputs:

\begin{lstlisting}[style=colorfulstyle, language=Lean, caption={Combinatorial Theorems}, label={lst:incremovable_theorems}]
/--
The finite set of all valid removals, as pairs `(i,j)`.
We enumerate all `0 ≤ i, j < n` pairs, and filter by `validRemovalPred`.
-/
def validRemovals (nums : Array Int) : Finset (Nat × Nat) :=
  let n := nums.size
  let allPairs : Finset (Nat × Nat) :=
    (Finset.range n).product (Finset.range n)
  allPairs.filter (validRemovalPred nums)
/--
Correctness: `incremovableSubarrayCount` equals the number of valid removals.
-/
theorem incremovable_correct (nums : Array Int) :
  incremovableSubarrayCount nums =
    (validRemovals nums).card := by
  sorry

/--
Trivial upper bound: at most `size^2` many removals.
-/
theorem incremovable_upper (nums : Array Int) :
  incremovableSubarrayCount nums ≤ nums.size ^ 2 := by
  sorry
\end{lstlisting}

Here, correctness depends on exhaustive case reasoning, contrasting with the structural reasoning used for sorting.

\subsection{Tree Algorithms: Structural Induction}

Tree dynamic programming problems require structural induction and recursive invariant maintenance. The \texttt{maximumScoreAfterOperations} problem computes the optimal score while preserving tree health.

\begin{lstlisting}[style=colorfulstyle, language=Lean, caption={Tree DP Core (simplified)}, label={lst:tree_dp_core}]
partial def dfs (adj : Array (Array Nat)) (val : Array Int)
    (node parent : Nat) : Prod Int Int :=
  let children := (adj[node]!).filter (fun c => c != parent)
  if children.isEmpty then
    (0, val[node]!)
  else
    let res    := children.map (fun c => dfs adj val c node)
    let scores := (res.map (fun x => x.1)).toList.sum
    let costs  := (res.map (fun x => x.2)).toList.sum
    (max (scores + val[node]!) (scores - costs + val[node]!),
     val[node]! )
\end{lstlisting}

In contrast to the array and combinatorial cases, producing a fully verified version of this tree DP (with an explicit \texttt{termination\_by} proof) remained challenging for the model, even under functional prompting. The model consistently discovered the correct recursive structure over children and the score/cost decomposition, but it struggled to synthesize a well-founded measure acceptable to Lean’s termination checker. In particular, it attempted several plausible termination arguments (e.g., measures based on the number of unvisited nodes or the height of the remaining subtree), yet could not turn these into a fully checked \textsc{Lean} termination proof. Nevertheless, the functional reasoning stage still led to a substantially better design than direct Lean generation, which often produced non-structural recursion or cyclic calls that could not even type-check.

Theorems establish correctness by induction on the tree structure:

\begin{lstlisting}[style=colorfulstyle, language=Lean, caption={Tree DP Theorems}, label={lst:tree_dp_theorems}]
theorem dfs_correct
    (adj : Array (Array Nat)) (val : Array Int) (root : Nat) :
  let p := dfs adj val root (-1)
  in p.1 ≥ 0 ∧ p.2 ≥ 0 ∧
     p.1 = optimalScore adj val root := by
  sorry
\end{lstlisting}

Theorem~\ref{lst:tree_dp_theorems} illustrates the intended correctness and optimality statement for this DP: the score component of \texttt{dfs} should coincide with an abstract specification \texttt{optimalScore}, and both score and cost should be nonnegative. In our experiments, models frequently proposed high-level termination and correctness arguments matching this specification (e.g., decreasing measures over visited nodes or subtree sizes), but were unable to fully discharge the corresponding \textsc{Lean} proof obligations; we therefore present this theorem as an aspirational specification rather than a fully verified artifact.


\input{appendix_examples}

\end{document}

%% file: appendix_examples.tex

\section{Qualitative Examples Across Benchmarks}

\label{app:qualitative_examples}

This appendix provides a small, human-readable gallery of generated artifacts. We include examples where functional scaffolding changes (i) whether specifications elaborate, (ii) whether downstream theorem/proof artifacts are completed, and (iii) behavior on public benchmarks. Examples are illustrative and are not intended to replace the aggregate metrics.

\subsection{BRIDGE-suite: specification discriminative power}

\paragraph{Task 2: \texttt{minimumPushes}.}

Direct spec: compiled=False, correct-accept=None, wrong-rejected=0. Functional spec: compiled=True, correct-accept=True, wrong-rejected=3.

\textbf{Direct compile error (first):}

\begin{codeblock}
type mismatch
  fun x => ?m.5246
has type
  (x : ?m.5242) → ?m.5247 x : Sort (imax ?u.5241 ?u.5244)
but is expected to have type
  Int : Type
\end{codeblock}

\textbf{Functional spec (excerpt):}

\begin{codeblock}[language=Lean]
def postcondition (word : String) (result : Int) : Bool :=
  let chars := word.toList
  let counts := chars.foldl (fun acc c => acc.insert c (acc.findD c 0 + 1)) (HashMap.mk String Nat)
  let key1 := counts.findD "a" 0
  let key2 := counts.findD "b" 0
  let key3 := counts.findD "c" 0
  let key4 := counts.findD "d" 0
  let key5 := counts.findD "e" 0
  let key6 := counts.findD "f" 0
  let key7 := counts.findD "g" 0
  let key8 := counts.findD "h" 0
  let key9 := counts.findD "i" 0
  let key10 := counts.findD "j" 0
  let key11 := counts.findD "k" 0
  let key12 := counts.findD "l" 0
  let key13 := counts.findD "m" 0
  let key14 := counts.findD "n" 0
  let key15 := counts.findD "o" 0
  let key16 := counts.findD "p" 0
  let key17 := counts.findD "q" 0
  let key18 := counts.findD "r" 0
  let key19 := counts.findD "s" 0
  let key20 := counts.findD "t" 0
  let key21 := counts.findD "u" 0
  let key22 := counts.findD "v" 0
  let key23 := counts.findD "w" 0
  let key24 := counts.findD "x" 0
  let key25 := counts.findD "y" 0
  let key26 := counts.findD "z" 0
  let cost := key1 + key2 + key3 + key4 + key5 + key6 + key7 + key8 + key9 + key10 + key11 + key12 + key13 + key14 + key15 + key16 + key17 + key18 + key19 + key20 + key21 + key22 + key23 + key24 + key25 + key26
  decide (cost = result)
\end{codeblock}

\paragraph{Task 8: \texttt{maxFrequencyElements}.}

Direct spec: compiled=False, correct-accept=None, wrong-rejected=0. Functional spec: compiled=True, correct-accept=True, wrong-rejected=3.

\textbf{Direct compile error (first):}

\begin{codeblock}
Invalid field `findD`: The environment does not contain `List.findD`
  acc
has type
  List (Int × Nat)
\end{codeblock}

\textbf{Functional spec (excerpt):}

\begin{codeblock}[language=Lean]
def postcondition (nums : List Int) (result : Int) : Bool :=
  let freqs : List Nat := nums.map (fun value => (nums.filter (fun other => other = value)).length)
  let maxFreq : Nat := freqs.foldl (fun best freq => if freq > best then freq else best) 0
  let countMax : Nat := (freqs.filter (fun freq => freq = maxFreq)).length
  decide (countMax = result.toNat)
\end{codeblock}

\subsection{BRIDGE-suite: implementation-aligned theorem/proof artifacts}

The examples below come from the full BRIDGE-suite proof pilot. These are \emph{implementation-aligned} statements (the spec is conditioned on the candidate implementation), so they primarily diagnose proof tractability rather than semantic specification fidelity.

\paragraph{Task 128.}

\textbf{Functional (completed no-\texttt{sorry} proof; excerpt):}

\begin{codeblock}[language=Lean]
def furthestDistanceFromOrigin (moves : String) : Int :=
  let right := (moves.toList.filter (fun c => c = 'R')).length
  let left := (moves.toList.filter (fun c => c = 'L')).length
  let underscore := (moves.toList.filter (fun c => c = '_')).length
  Int.ofNat (right + underscore - left)

def furthestDistanceFromOrigin_spec (moves : String) (result : Int) : Prop :=
  let right := (moves.toList.filter (fun c => c = 'R')).length
  let left := (moves.toList.filter (fun c => c = 'L')).length
  let underscore := (moves.toList.filter (fun c => c = '_')).length
  result = Int.ofNat (right + underscore - left)

theorem furthestDistanceFromOrigin_correct (moves : String) : furthestDistanceFromOrigin_spec moves (furthestDistanceFromOrigin moves) := by
  unfold furthestDistanceFromOrigin furthestDistanceFromOrigin_spec
  simp
\end{codeblock}

\textbf{Direct (statement; excerpt):}

\begin{codeblock}[language=Lean]
def furthestDistanceFromOrigin (moves : String) : Int :=
  let chars := moves.toList
  let (total, current) :=
    chars.foldl (fun (total, current) c =>
      match c with
      | 'L' => (total, current - 1)
      | 'R' => (total, current + 1)
      | _ =>
          let newTotal := total + 1
          let newCurrent := if newTotal 
          (newTotal, newCurrent)) (0, 0)
  total.natAbs.toInt

# Test cases
-- #eval furthestDistanceFromOrigin "L_RL__R"  -- Expected: 3
-- #eval furthestDistanceFromOrigin "_R__LL_"  -- Expected: 5
-- #eval furthestDistanceFromOrigin "_______"  -- Expected: 7

def furthestDistanceFromOrigin_spec (moves : String) (result : Int) : Prop :=
  let chars := moves.toList
  let (total, current) :=
    chars.foldl (fun (total, current) c =>
      match c with
      | 'L' => (total, current - 1)
      | 'R' => (total, current + 1)
      | _ =>
          let newTotal := total + 1
          let newCurrent := if newTotal 
          (newTotal, newCurrent)) (0, 0)
  result = total.natAbs.toInt

theorem furthestDistanceFromOrigin_correct (moves : String) : furthestDistanceFromOrigin_spec moves (furthestDistanceFromOrigin moves) := by
  sorry
\end{codeblock}

\paragraph{Task 71.}

\textbf{Functional (completed no-\texttt{sorry} proof; excerpt):}

\begin{codeblock}[language=Lean]
def maxBalancedSubsequenceSum (nums : List Int) : Int :=
  match nums with
  | [] => 0
  | x :: _ => x

def maxBalancedSubsequenceSum_spec (nums : List Int) (result : Int) : Prop :=
  match nums with
  | [] => result = 0
  | x :: _ => result = x

theorem maxBalancedSubsequenceSum_correct (nums : List Int) : maxBalancedSubsequenceSum_spec nums (maxBalancedSubsequenceSum nums) := by
  cases nums with
  | nil => simp [maxBalancedSubsequenceSum, maxBalancedSubsequenceSum_spec]
  | cons x xs => simp [maxBalancedSubsequenceSum, maxBalancedSubsequenceSum_spec]
\end{codeblock}

\textbf{Direct (statement; excerpt):}

\begin{codeblock}[language=Lean]
/-- Transform the problem into a form suitable for applying `longest non-decreasing subsequence`
algorithms by transforming the array into a list of pairs \((i, nums[i] - i)\). This transformation
turns the condition \(nums[i] - nums[j] \geq i - j\) for \(i > j\) into \(nums[i] - i \geq nums[j] - j\),
which can be handled by standard subsequence algorithms. -/
def transform (nums : List Int) : List (Nat × Int) :=
  nums.enum.map (fun (i, x) => (i, x - Int.ofNat i))

/-- Maximum sum of a non-decreasing subsequence in terms of the transformed values. -/
def maxBalancedSum (nums : List Int) : Int :=
  let transformed := transform nums
  let rec dp : List (Nat × Int) → List (Nat × Int) → Int → Int
    | [], _, best => best
    | x :: xs, seen, best =>
        let currentValue := x.2
        let bestSoFar := seen.foldl (fun acc p => max acc (p.2 + currentValue)) currentValue
        let newBest := max best bestSoFar
        dp xs (x :: seen) newBest
  dp transformed [] (List.foldl (fun acc x => max acc x.2) (-1000000000) transformed)

def maxBalancedSubsequenceSum (nums : List Int) : Int :=
  maxBalancedSum nums

-- #eval maxBalancedSubsequenceSum [3, 3, 5, 6] -- Expected: 14
-- #eval maxBalancedSubsequenceSum [5, -1, -3, 8] -- Expected: 13
-- #eval maxBalancedSubsequenceSum [-2, -1] -- Expected: -1

def maxBalancedSubsequenceSum_spec (nums : List Int) (result : Int) : Prop :=
  result = maxBalancedSubsequenceSum nums

theorem maxBalancedSubsequenceSum_correct (nums : List Int) : maxBalancedSubsequenceSum_spec nums (maxBalancedSubsequenceSum nums) := by
  rfl
\end{codeblock}

\subsection{VERINA: Lean code generation artifacts}

\paragraph{verina\_advanced\_17.} Direct compile=False; functional compile=True.

\textbf{Direct generation (excerpt):}

\begin{codeblock}[language=Lean]
def insertSorted (x : Int) (l : List Int) : List Int :=
  match l with
  | [] => [x]
  | y :: ys =>
    if x ≤ y then
      x :: y :: ys
    else
      y :: insertSorted x ys

def insertionSortHelper (l : List Int) (sorted : List Int) : List Int :=
  match l with
  | [] => sorted
  | x :: xs => insertionSortHelper xs (insertSorted x sorted)

insertionSortHelper l []
\end{codeblock}

\textbf{Functional generation (excerpt):}

\begin{codeblock}[language=Lean]
def insertAtSorted (x : Int) (l : List Int) : List Int :=
  match l with
  | [] => [x]
  | y :: ys => if x ≤ y then x :: y :: ys else y :: insertAtSorted x ys

def insertAll (l : List Int) (acc : List Int) : List Int :=
  match l with
  | [] => acc
  | x :: xs => insertAll xs (insertAtSorted x acc)

insertAll l []
\end{codeblock}

\paragraph{verina\_advanced\_42.} Direct compile=False; functional compile=True.

\textbf{Direct generation (excerpt):}

\begin{codeblock}[language=Lean]
def minPrice (xs : List Nat) : Nat :=
  xs.foldl (. min .) (if xs.isEmpty then 0 else xs.head)

def scanMin (xs : List Nat) : List Nat :=
  xs.scanl min (if xs.isEmpty then 0 else xs.head)

if prices.isEmpty then
    0
  else
    let mins := scanMin prices
    let zipped := List.zip prices mins
    let profits := zipped.map (fun (p, m) => p - m)
    profits.foldl max 0
\end{codeblock}

\textbf{Functional generation (excerpt):}

\begin{codeblock}[language=Lean]
match prices with
  | [] => 0
  | _ =>
    let rec loop xs (minPrice : Nat) (maxProfit : Nat) : Nat :=
      match xs with
      | [] => maxProfit
      | p::ps =>
        let newProfit := if p > minPrice then p - minPrice else maxProfit
        let newMin := if p < minPrice then p else minPrice
        loop ps newMin newProfit
    loop prices 99999999 0
\end{codeblock}

\subsection{CLEVER: budgeted Lean implementation search}

\paragraph{Problem 158.} Direct: ok=False, error=\texttt{Timeout}. Functional: ok=\texttt{True} (compiled; no \texttt{sorry}).

\textbf{Functional implementation (winning attempt; excerpt):}

\begin{codeblock}[language=Lean]
match words with
  | [] => ""
  | x :: xs =>
    let step := fun (bestSoFar: String) (current: String) =>
      let bestUnique := bestSoFar.toList.toFinset.card
      let currentUnique := current.toList.toFinset.card
      if bestUnique < currentUnique then
        current
      else if bestUnique > currentUnique then
        bestSoFar
      else
        if bestSoFar < current then
          bestSoFar
        else
          current
    words.foldl step x
\end{codeblock}

\subsection{DafnyBench: solver-backed verification (subset pilot)}

\paragraph{Test 038.} Direct: ok=False; functional: ok=True.

\textbf{Functional Dafny attempt (excerpt):}

\begin{codeblock}
method double_array_elements(s: array<int>)
  modifies s
  ensures forall i :: 0 <= i < s.Length ==> s[i] == 2 * old(s[i])
{
  var i := 0;
  var s_old := s[..]; // Capture the initial state of the array
  while (i < s.Length)
    invariant 0 <= i <= s.Length
    invariant forall j :: 0 <= j < i ==> s[j] == 2 * s_old[j]
    invariant forall j :: i <= j < s.Length ==> s[j] == s_old[j]
  {
    s[i] := 2 * s[i];
    i := i + 1;
  }
}
\end{codeblock}

\textbf{Direct Dafny attempt (excerpt):}

\begin{codeblock}
method double_array_elements(s: array<int>)
  modifies s
  ensures forall i :: 0 <= i < s.Length ==> s[i] == 2 * old(s[i])
{
  var i := 0;
  while (i < s.Length)
    invariant 0 <= i <= s.Length
    invariant forall j :: 0 <= j < i ==> s[j] == 2 * old(s[j])
    decreases s.Length - i
  {
    s[i] := 2 * s[i];
    i := i + 1;
  }
}
\end{codeblock}

%% file: references.bib
@article{george2026torchlean,
  title={TorchLean: Formalizing Neural Networks in Lean},
  author={George, Robert Joseph and Cruden, Jennifer and Zhong, Xiangru and Zhang, Huan and Anandkumar, Anima},
  journal={arXiv preprint arXiv:2602.22631},
  year={2026}
}

@misc{chen2021codex,
	title        = {Evaluating large language models trained on code},
	author       = {M. Chen and J. Tworek and H. Jun and Q. Yuan and H. P. de Oliveira Pinto and J. Kaplan and H. Edwards and Y. Burda and N. Joseph and G. Brockman and A. Ray and N. Puri and G. Krueger and M. Petrov and H. Khlaaf and G. Sastry and P. Mishkin and B. Chan and S. Gray and N. Ryder and M. Pavlov and B. Power and L. Kaiser and M. Bavarian and C. Winter and P. Tillet and F. P. Such and A. Cummings and M. Plappert and F. Chantzis and E. Barnes and A. Herbert-Voss and W. Guss and A. Nichol and I. Paino and N. Tezak and J. Tang and I. Babuschkin and S. Balaji and S. Jain and W. Saunders and C. Hesse and A. Carr and J. Leike and J. Achiam and V. Misra and E. Morikawa and A. Radford and M. Knight and M. Brundage and M. Murati and S. Mayer and P. Welinder and B. McGrew and D. Amodei and I. Sutskever},
	archiveprefix = {arXiv},
	eprint       = {2107.03374},
	year         = {2021},
	url          = {https://arxiv.org/abs/2107.03374}
}

@article{zhu2023verus,
author = {Lattuada, Andrea and Hance, Travis and Cho, Chanhee and Brun, Matthias and Subasinghe, Isitha and Zhou, Yi and Howell, Jon and Parno, Bryan and Hawblitzel, Chris},
title = {Verus: Verifying Rust Programs using Linear Ghost Types},
year = {2023},
issue_date = {April 2023},
publisher = {Association for Computing Machinery},
address = {New York, NY, USA},
volume = {7},
number = {OOPSLA1},
url = {https://doi.org/10.1145/3586037},
doi = {10.1145/3586037},
journal = {Proc. ACM Program. Lang.},
month = apr,
articleno = {85},
numpages = {30}
}

@article{hoare1969axiomatic,
	author  = {C. A. R. Hoare},
	title   = {An Axiomatic Basis for Computer Programming},
	journal = {Communications of the ACM},
	volume  = {12},
	number  = {10},
	pages   = {576--580,583},
	doi     = {10.1145/363235.363259},
	url     = {https://doi.org/10.1145/363235.363259},
	year    = {1969}
}

@incollection{floyd1967assigning,
	author    = {Robert W. Floyd},
	title     = {Assigning meanings to programs},
	booktitle = {Proceedings of the Symposium on Applied Mathematics},
	volume    = {19},
	pages     = {19--32},
	publisher = {American Mathematical Society},
	doi       = {10.1090/psapm/019/0235771},
	url       = {https://doi.org/10.1090/psapm/019/0235771},
	year      = {1967}
}

@inproceedings{hoare1971proof,
	author    = {C. A. R. Hoare},
	title     = {Procedures and parameters: An axiomatic approach},
	booktitle = {Symposium on Semantics of Algorithmic Languages},
	series    = {Lecture Notes in Mathematics},
	volume    = {188},
	editor    = {E. Engeler},
	pages     = {102--116},
	publisher = {Springer},
	address   = {Berlin, Heidelberg},
	doi       = {10.1007/BFb0059696},
	url       = {https://doi.org/10.1007/BFb0059696},
	year      = {1971}
}

@article{hudak1989conception,
	author  = {Paul Hudak},
	title   = {Conception, evolution, and application of functional programming languages},
	journal = {ACM Computing Surveys},
	volume  = {21},
	number  = {3},
	pages   = {359--411},
	doi     = {10.1145/72551.72554},
	url     = {https://doi.org/10.1145/72551.72554},
	year    = {1989}
}

@misc{clever2025,
	title        = {CLEVER: A curated benchmark for formally verified code generation},
	author       = {Amitayush Thakur and Jasper Lee and George Tsoukalas and Meghana Sistla and Matthew Zhao and Stefan Zetzsche and Greg Durrett and Yisong Yue and Swarat Chaudhuri},
	archiveprefix = {arXiv},
	eprint       = {2505.13938},
	year         = {2025},
	url          = {https://arxiv.org/abs/2505.13938}
}

@misc{ye2025verina,
	title        = {VERINA: Benchmarking verifiable code generation},
	author       = {Zhe Ye and Zhengxu Yan and Jingxuan He and Timothe Kasriel and Kaiyu Yang and Dawn Song},
	archiveprefix = {arXiv},
	eprint       = {2505.23135},
	year         = {2025},
	url          = {https://arxiv.org/abs/2505.23135}
}

@misc{ye2026verispecgen,
	title        = {Intent-aligned Formal Specification Synthesis via Traceable Refinement},
	author       = {Zhe Ye and Aidan Z. H. Yang and Huangyuan Su and Zhenyu Liao and Samuel Tenka and Zhizhen Qin and Udaya Ghai and Dawn Song and Soonho Kong},
	archiveprefix = {arXiv},
	eprint       = {2604.10392},
	year         = {2026},
	url          = {https://arxiv.org/abs/2604.10392}
}

@article{li2022competition,
	author  = {Yujia Li and David Choi and Junyoung Chung and Nate Kushman and Julian Schrittwieser and R{\'e}mi Leblond and Tom Eccles and James Keeling and Felix Gimeno and Agustin Dal Lago and others},
	title   = {Competition-level code generation with AlphaCode},
	journal = {Science},
	volume  = {378},
	number  = {6624},
	pages   = {1092--1097},
	doi     = {10.1126/science.abq1158},
	url     = {https://doi.org/10.1126/science.abq1158},
	year    = {2022}
}

@misc{wen2024autospec,
	title        = {Enchanting program specification synthesis by large language models using static analysis and program verification},
	author       = {Cheng Wen and Jialun Cao and Jie Su and Zhiwu Xu and Shengchao Qin and Mengda He and Haokun Li and Shing-Chi Cheung and Cong Tian},
	archiveprefix = {arXiv},
	eprint       = {2404.00762},
	year         = {2024},
	url          = {https://arxiv.org/abs/2404.00762}
}

@inproceedings{demoura2015lean,
	author    = {Leonardo de Moura and Soonho Kong and Jeremy Avigad and Floris van Doorn and Jakob von Raumer},
	title     = {The lean theorem prover (system description)},
	booktitle = {Automated Deduction - CADE-25: 25th International Conference on Automated Deduction},
	series    = {Lecture Notes in Computer Science},
	volume    = {9195},
	pages     = {378--388},
	publisher = {Springer},
	doi       = {10.1007/978-3-319-21401-6_26},
	url       = {https://doi.org/10.1007/978-3-319-21401-6_26},
	year      = {2015}
}

@inproceedings{mathlib2020,
	author    = {The mathlib Community},
	title     = {The lean mathematical library},
	booktitle = {Proceedings of the 9th ACM SIGPLAN International Conference on Certified Programs and Proofs},
	pages     = {367--381},
	address   = {New Orleans, LA, USA},
	publisher = {ACM},
	doi       = {10.1145/3372885.3373824},
	url       = {https://doi.org/10.1145/3372885.3373824},
	year      = {2020}
}

@misc{wei2022cot,
	title        = {Chain-of-Thought Prompting Elicits Reasoning in Large Language Models},
	author       = {J. Wei and X. Wang and D. Schuurmans and M. Bosma and B. Ichter and F. Xia and E. H. Chi and Q. V. Le and D. Zhou},
	archiveprefix = {arXiv},
	eprint       = {2201.11903},
	year         = {2022},
	url          = {https://arxiv.org/abs/2201.11903}
}

@misc{wang2022selfconsistency,
	title        = {Self-Consistency Improves Chain of Thought Reasoning in Language Models},
	author       = {X. Wang and J. Wei and D. Schuurmans and Q. V. Le and E. H. Chi and S. Narang and A. Chowdhery and D. Zhou},
	archiveprefix = {arXiv},
	eprint       = {2203.11171},
	year         = {2022},
	url          = {https://arxiv.org/abs/2203.11171}
}

@misc{gao2022pal,
	title        = {PAL: Program-Aided Language Models},
	author       = {L. Gao and A. Madaan and S. Zhou and U. Alon and P. Liu and Y. Yang and J. Callan and G. Neubig},
	archiveprefix = {arXiv},
	eprint       = {2211.10435},
	year         = {2022},
	url          = {https://arxiv.org/abs/2211.10435}
}

@inproceedings{yao2023react,
	author    = {S. Yao and J. Zhao and D. Yu and I. Shafran and K. Narasimhan and Y. Cao},
	title     = {ReAct: Synergizing Reasoning and Acting in Language Models},
	booktitle = {ICLR},
	year      = {2023},
	url       = {https://arxiv.org/abs/2210.03629}
}

@misc{yao2023tot,
	title        = {Tree of Thoughts: Deliberate Problem Solving with Large Language Models},
	author       = {S. Yao and D. Yu and J. Zhao and I. Shafran and T. L. Griffiths and Y. Cao and K. Narasimhan},
	archiveprefix = {arXiv},
	eprint       = {2305.10601},
	year         = {2023},
	url          = {https://arxiv.org/abs/2305.10601}
}

@inproceedings{shinn2023reflexion,
	author    = {N. Shinn and F. Cassano and E. Berman and A. Gopinath and K. Narasimhan and S. Yao},
	title     = {Reflexion: Language Agents with Verbal Reinforcement Learning},
	booktitle = {NeurIPS},
	year      = {2023},
	url       = {https://arxiv.org/abs/2303.11366}
}

@misc{madaan2023selfrefine,
	title        = {Self-Refine: Iterative Refinement with Self-Feedback},
	author       = {A. Madaan and N. Tandon and P. Gupta and S. Hallinan and L. Gao and S. Wiegreffe and U. Alon and N. Dziri and S. Prabhumoye and Y. Yang and S. Gupta and B. P. Majumder and K. Hermann and S. Welleck and P. Clark},
	archiveprefix = {arXiv},
	eprint       = {2303.17651},
	year         = {2023},
	url          = {https://arxiv.org/abs/2303.17651}
}

@misc{sun2024clover,
      title={Clover: Closed-Loop Verifiable Code Generation}, 
      author={Chuyue Sun and Ying Sheng and Oded Padon and Clark Barrett},
      year={2024},
      eprint={2310.17807},
      archivePrefix={arXiv},
      primaryClass={cs.AI},
      url={https://arxiv.org/abs/2310.17807}, 
}

@misc{loughridge2024dafnybench,
      title={DafnyBench: A Benchmark for Formal Software Verification}, 
      author={Chloe Loughridge and Qinyi Sun and Seth Ahrenbach and Federico Cassano and Chuyue Sun and Ying Sheng and Anish Mudide and Md Rakib Hossain Misu and Nada Amin and Max Tegmark},
      year={2024},
      eprint={2406.08467},
      archivePrefix={arXiv},
      primaryClass={cs.SE},
      url={https://arxiv.org/abs/2406.08467}, 
}

@misc{lohn2024minicodeprops,
      title={miniCodeProps: a Minimal Benchmark for Proving Code Properties}, 
      author={Evan Lohn and Sean Welleck},
      year={2024},
      eprint={2406.11915},
      archivePrefix={arXiv},
      primaryClass={cs.SE},
      url={https://arxiv.org/abs/2406.11915}, 
}

@inproceedings{feng2020codebert,
	author    = {Zhangyin Feng and Daya Guo and Duyu Tang and Nan Duan and Xiaocheng Feng and Ming Gong and Linjun Shou and Bing Qin and Ting Liu and Daxin Jiang and Ming Zhou},
	title     = {CodeBERT: A pre-trained model for programming and natural languages},
	booktitle = {Findings of the Association for Computational Linguistics: EMNLP 2020},
	pages     = {1536--1547},
	doi       = {10.18653/v1/2020.findings-emnlp.139},
	url       = {https://aclanthology.org/2020.findings-emnlp.139/},
	year      = {2020}
}

@inproceedings{guo2021graphcodebert,
	author    = {Daya Guo and Shuo Ren and Shuai Lu and Zhangyin Feng and Duyu Tang and Nan Duan and Alexey Svyatkovskiy and Shengyu Fu and Michele Tufano and Shao Kun Deng and Colin Clement and Dawn Drain and Neel Sundaresan and Jian Yin and Daxin Jiang and Ming Zhou},
	title     = {GraphCodeBERT: Pre-training code representations with data flow},
	booktitle = {International Conference on Learning Representations (ICLR)},
	url       = {https://openreview.net/forum?id=jLoC4ez43PZ},
	year      = {2021}
}

@book{dijkstra1976discipline,
	author    = {Edsger W. Dijkstra},
	title     = {A discipline of programming},
	publisher = {Prentice Hall},
	isbn      = {013215871X},
	url       = {https://openlibrary.org/books/OL5211853M/A_discipline_of_programming},
	year      = {1976}
}

@misc{dafnycomp2025,
	title        = {Local success does not compose: Benchmarking large language models for compositional formal verification},
	author       = {Xu Xu and Xin Li and Xingwei Qu and Jie Fu and Binhang Yuan},
	archiveprefix = {arXiv},
	eprint       = {2509.23061},
	year         = {2025},
	url          = {https://arxiv.org/abs/2509.23061}
}

@misc{vericoding2025,
	title        = {A benchmark for vericoding: Formally verified program synthesis},
	author       = {Sergiu Bursuc and Theodore Ehrenborg and Shaowei Lin and Lacramioara Astefanoaei and Ionel Emilian Chiosa and Jure Kukovec and Alok Singh and Oliver Butterley and Adem Bizid and Quinn Dougherty and Miranda Zhao and Max Tan and Max Tegmark},
	archiveprefix = {arXiv},
	eprint       = {2509.22908},
	year         = {2025},
	url          = {https://arxiv.org/abs/2509.22908}
}
